\documentclass{article}

\usepackage{microtype}
\usepackage{graphicx}
\usepackage{subcaption}
\usepackage{booktabs} 
\usepackage{enumitem}
\usepackage{threeparttable}

\usepackage[table,x11names,dvipsnames,table,usenames]{xcolor}
\usepackage{stfloats}
\usepackage[export]{adjustbox}
\usepackage{tablefootnote}
\usepackage{newfloat}
\usepackage{listings}
\usepackage{multirow}

\usepackage[most]{tcolorbox}
\usepackage{xcolor}
\usepackage{fvextra}

\definecolor{colorPrompt}{HTML}{009E73}

\DeclareCaptionType[within=none]{promptbox}[Prompt][List of prompts]

\newtcolorbox{prompt}[4][]{
  breakable,
  colback=colorPrompt!5!white,
  colframe=colorPrompt!75!black,
  fonttitle=\bfseries\small,
  fontupper=\small,
  title={#2},
  after={\captionsetup{type=promptbox}\captionof{promptbox}{#3}\label{#4}\vspace{3mm}},
  #1
}

\DefineVerbatimEnvironment{promptjson}{Verbatim}{
  breaklines=true,
  breakanywhere=true,
  fontsize=\scriptsize,
  baselinestretch=0.95
}

\usepackage{pgfplots}
\usepackage{pgfplotstable}
\pgfplotsset{compat=1.18}
\usepgfplotslibrary{groupplots} 
\usetikzlibrary{shapes.geometric, arrows.meta, positioning, calc, fit, shadows}
\usepackage[outline]{contour} 
\contourlength{1.5pt}

\usepackage{wrapfig}
\usepackage{placeins}

\usepackage[numbers,sort&compress]{natbib}

\usepackage{hyperref}
\lstdefinestyle{promptstyle}{
  basicstyle=\ttfamily\small,
  breaklines=true,
  columns=fullflexible,
  frame=single,
  rulecolor=\color{black!20},
  showstringspaces=false
}

\usepackage{amsmath}
\usepackage{amssymb}
\usepackage{mathtools}
\usepackage{amsthm}

\usepackage[capitalize,noabbrev]{cleveref}
\Crefname{section}{Sec.}{Secs.}
\Crefname{figure}{Fig.}{Figs.}
\Crefname{table}{Tab.}{Tabs.}

\theoremstyle{plain}

\theoremstyle{definition}

\theoremstyle{remark}

\usepackage[normalem]{ulem}

\usepackage[disable,textsize=tiny]{todonotes}
\usepackage[textsize=tiny]{todonotes}
\usepackage{pifont}

\newcommand\TODO[1]{\textbf{\textcolor{red}{[TODO: #1]}}}

\newcommand{\ie}[0]{\textit{i.e.}}

\newcommand{\repr}{\text{reproducibility}}
\newcommand{\rep}[1]{\text{{#1}eproducibility}}

\definecolor{MarkGreen}{RGB}{0, 180, 0}
\definecolor{MarkRed}{RGB}{248, 3, 7} 
\definecolor{MarkYellow}{RGB}{160, 160, 0}
\definecolor{MarkBrown}{RGB}{191, 135, 93}
\newcommand{\cmark}{{\textcolor{MarkGreen}{\ding{51}}}}%
\newcommand{\qmark}{{\textcolor{YellowOrange}{\textbf{?}}}}%
\newcommand{\xmark}{{\textcolor{MarkRed}{\ding{55}}}}%
\newcommand{\nmark}{{\textcolor{MarkBrown}{$\pmb\varnothing$}}}%

\definecolor{cCVPR}{RGB}{0,114,178}   
\definecolor{cICCV}{RGB}{213,94,0}    
\definecolor{cICLR}{RGB}{0,158,115}   
\definecolor{cICML}{RGB}{204,121,167} 
\definecolor{cNEUR}{RGB}{230,159,0}   
\definecolor{cALL}{RGB}{86,180,233}   

\definecolor{topic1}{RGB}{0,114,178}    
\definecolor{topic2}{RGB}{213,94,0}     
\definecolor{topic3}{RGB}{0,158,115}    
\definecolor{topic4}{RGB}{204,121,167}  
\definecolor{topic5}{RGB}{230,159,0}    
\definecolor{topic6}{RGB}{86,180,233}   
\definecolor{topic7}{RGB}{240,228,66}   
\definecolor{topic8}{RGB}{0,0,0}        
\definecolor{topic9}{RGB}{117,112,179}  
\definecolor{topic10}{RGB}{102,166,30}  

\newcommand\field[1]{\textsl{#1}}
\def\App{App.}
\newcommand{\appref}[1]{\App~\ref{#1}}

\makeatletter

\definecolor{bgGreen}{RGB}{213,232,212}
\definecolor{bgRed}{RGB}{248,206,204}
\definecolor{bgYellow}{RGB}{255,242,204}
\definecolor{bgOrange}{RGB}{255,230,204}

\usepackage{url}

\newcommand{\boldparagraph}[1]{\vspace{0.01cm}\noindent{\bf #1.} \vspace{0.01cm}}

\usepackage[nonatbib,position, final]{neurips_2026}

\title{Position: Let’s Strengthen Verifiability \\ If We Can't Enforce Reproducibility}

\author{%
  Samet Hicsonmez \\
  Independent Researcher \\
  Paris, France \\
  \And
  Nermin Samet \\
  Valeo.ai \\
  Paris, France \\
  \And
  Renaud Marlet \\
  Valeo.ai \\
  Paris, France \\
}

\begin{document}

\maketitle

\begin{abstract}

In the field of Machine Learning, many papers contain empirical results supporting claimed statements or illustrating the performance of a proposed method. However, most practitioners know that (1)~results are generally hard to reproduce, and increasingly so, (2)~code is not often available to do so, and (3)~it hinders the development of research. In this position paper, we analyze and quantify these issues, and make concrete proposals to improve result checkability, if not reproducibility. Code and supporting materials are available at https://github.com/giddyyupp/position-enforce-verifiability.
\end{abstract}

\section{Introduction}

Reproducibility is the cornerstone of scientific progress. This foundational statement \cite{popper1934}, likely as old as epistemology, with figures as Plato and Aristotle, has been recalled many times in the last decades \cite{moonesinghe2007replication, simons2014replication}, especially since the replication crisis gained momentum about ten years ago \cite{phashler2012replicabilitycrisis, baker2016reproducibility,fanelli2018realcrisis,antunes2024reproducibilityhpc}. 

The fact is that the figures on failures to reproduce an experiment, whether one's own or somebody else's, are compelling: according to a study involving about 1,600 researchers in various fields of science \cite{baker2016reproducibility}, more than 70\% of them have tried and failed to reproduce another scientist’s experiments, and more than half have failed to reproduce one of their own experiments. 
In contrast, a recent study on AI/ML publications~\cite{nag2025global} found a low 0.05\% retraction rate, with papers that often continue to be cited, sometimes receiving up to 8 times more citations after retraction.

Several factors amplify this crisis: each year there are more researchers, more papers per researcher, and papers are more complex, harder to write and review.
Studies show indeed that, between 2014 and 2018, the number of researchers worldwide (approaching 9\,M) has increased by 3.3\% per year, 
which is 3 times faster than the global population \cite{unesco2021sciencereport}.
Also, from 2016 to 2022, the number of scientific publications worldwide (currently over 3\,M/year) has grown by 5.6\% per year \cite{hanson2024publishing}.
Because of the \emph{burden of knowledge} \cite{jones2009burdenofknowledge}, as papers require more expertise to write and review, reproducibility may also be hindered. Our experience is that it is increasingly difficult to publish a simple method, even if it outperforms the state of the art (SOTA), because then of an alleged ``lack of technical novelty''; this encourages authors to propose artificially complex methods, which are then also harder to reproduce. 

Last, Generative AI (GenAI) now automates the production of scientific contents, with possible errors or misuses. There are numerous reports of dishonest papers, partly or totally generated by AI, e.g., with hallucinated references \cite{halluciteneurips}, copycats of genuine papers \cite{naddaf2025copycat}, or paper mills \cite{cardenuto2024unveiling} with fake papers to boost citations \cite{alsinani2025contentcreationcitationinflation},
pushing arXiv to now require first-time posters to be endorsed \cite{boboris2026arxivpolicy}.

Compared to social, life, and physical sciences, computer science is fortunate to have straightforward assets to ensure reproducibility: code and data. Putting aside variations due to stochastic algorithms, fixed-precision computation, and residual randomness in modern hardware, which however are prevalent in Machine Learning (ML), computer science is assumed to be an exact science involving deterministic machines. In theory, reproduction can be confirmed by a simple keystroke. In practice, there are many barriers, and the reproducibility crisis affects computer science too, including ML \cite{antunes2024reproducibilityhpc}.

In this position paper, we study code availability (\cref{sec:codenotavail}) and its status regarding replication (\cref{sec:barriers}). Considering top-tier ML and Computer Vision (CV) conferences, we observe a recent decrease in the ratio of accepted papers with accessible code, though papers with code have twice as many citations on average. \textbf{Considering that reproducibility and code availability will remain hard to generalize, our position is that the community should at least improve the verifiability of experiments.} We therefore propose two verifications (\cref{sec:checking}) and an associated publication process (\cref{sec:policy}) that can contribute to improving the confidence in reported experiments.

\section{Related analyses and proposals}
\label{sec:related}

A broad consensus has emerged regarding \rep{r}{} in ML~\cite{pineau2021improving, raff2025machine}. Following ACM terminology, \repr{} asks: “Can an independent team obtain the reported results using the code and data provided by the original authors?” Under this definition, \repr{} assumes (i)~public access to the relevant artifacts, and (ii)~third-party verification of the reported outcomes. While access to code and data is necessary, some argue that this artifact release alone does not guarantee \repr{}~\cite{drummond2009replicability, shehzad2025we}. 
The more prominent obstacle however remains that code or data are often not released, motivating efforts to identify what makes a paper reproducible beyond code availability.

\citet{raff2019step} provides a concrete illustration: he attempted to reimplement 255 papers without consulting official code, even when available. Only 63.5\% of the papers could be replicated under the reproducibility criterion, requiring that at least 75\% of the claims be validated. A key finding is that readability, meaning clear implementation details and informative pseudocode, is the strongest predictor of \rep{r}{}. Furthermore, when the authors of the original paper help, the \rep{r}{} increases to 85\%.
More recently, Paper2Code~\cite{seo2025paper2code} explores whether \rep{r}{} can be automated by generating code from the paper alone. This LLM-based multi-agent system achieves an overall replication rate of roughly 45\% on a public benchmark~\cite{staracepaperbench}.

Several other works~\cite{haibe2020transparency, kapoor2023leakage, ravi2025improving, d2025lessons} investigated the prominent issues hindering \rep{r}{} in broader ML-based methods across various fields. 

This ever growing \rep{r}{} concern in different domains urged the community to take action.  
NeurIPS 2019 has introduced a ``\rep{R}{} Challenge" and an ``ML reproducibility checklist" to encourage and improve the \rep{r}{} of accepted papers \cite{pineau2021improving}. The former aims to provide \textit{independent} verification and validation of the empirical claims in accepted papers. (Only 173 papers were submitted for this challenge out of 1,427 accepted papers.) The latter involves responding to a questionnaire assessing whether the paper includes various essential information to ensure \rep{r}{}; it has to be completed during the initial submission phase. 

The Journal of Artificial Intelligence Research (JAIR)~\cite{gundersen2024improving} introduces four novel mechanisms to make AI research more reproducible: (1)~``\rep{r}{} checklists'', similar to the one mentioned above, (2)~``structured abstracts'' to present key components of the paper in a more structured way, (3)~``\rep{r}{} badges" to incentivize transparency and reproducibility by making data, code, both, or independent reproductions available, and finally (4)~``\rep{r}{} reports" to encourage independent \rep{r}{} of JAIR papers. More specifically, JAIR created an additional track for papers that present a public implementation and validation of a previously accepted JAIR paper.  

EuroSys~\cite{d2025lessons} proposes various short- and long-term directions to overcome the \rep{r}{} challenges after carefully analyzing artifact evaluation processes in previous editions. These recommendations include artifact submission at various stages of the reviewing process.

\section{Code availability study}
\label{sec:codenotavail}

Software (code and data) is central to reproducibility for computer science papers. In this section, we first consider general issues related to software availability (\cref{sec:issues_code_availability}), then study access to code of published papers in five top-tier ML and CV conferences over the last five years (\cref{sec:studycode}).

\subsection{Issues with software availability}
\label{sec:issues_code_availability}

\paragraph{Various degrees of code accessibility.}

The specificity of ML is that there are two computing stages: model learning form training data, and model inference, i.e., testing. It results in three common cases of code availability: (1)~no code available; (2)~trained model available, i.e., inference code and learned parameters (e.g., network weights) but not training code; (3)~both training and inference code available, possibly with some learned parameters. Still, it is not uncommon that, due to partial code availability, only a fraction of the experiments can be reproduced, e.g., not on all datasets, not with test-time augmentations, or not with ensembling.

\begin{wrapfigure}{r}{0.5\textwidth}
\centering
\vspace{-5ex}
\resizebox{\linewidth}{!}{


\begin{tikzpicture}
\begin{groupplot}[
  group style={
    group size=1 by 3,        
    vertical sep=2.1cm,
    ylabels at=edge left,
    yticklabels at=edge left,
  },
  width=\textwidth,
  height=0.35\textwidth,      
  symbolic x coords={2021,2022,2023,2024,2025},
  xtick=data,
  grid=both,
  grid style={black!10},
  scaled y ticks=false,
  yticklabel style={/pgf/number format/fixed, /pgf/number format/precision=2},
  ylabel={Ratio (\%)},              
  tick label style={font=\Large},
label style={font=\Large},        
title style={font=\Large},
  legend style={
    at={(0.5,-0.32)},
    anchor=north,
    draw=black!15,
    fill=white,
    fill opacity=0.9,
    rounded corners=2pt,
    font=\Large,
  },
  legend columns=6,
  clip=false,
  cycle list={
      {color=cCVPR, very thick, mark=o},
      {color=cICCV, very thick, mark=square*},
      {color=cICLR, very thick, mark=triangle*},
      {color=cICML, very thick, mark=diamond*},
      {color=cNEUR, very thick, mark=pentagon*},
      {color=cALL,  dashed, very thick, mark=*},
  },
  enlarge x limits=0.03,
]
\pgfplotstableread{
Year  CVPR  ICCV  ICLR  ICML  NeurIPS  ALL
2021  39.1  42.1  47.7  31.3  38.4     39.3
2022  47.7  nan   52.2  40.1  38.2     43.4
2023  45.7  47.9  52.0  41.8  46.3     46.8
2024  46.2  nan   52.2  43.7  50.9     48.7
2025  44.1  41.0  51.3  44.5  49.3     47.1
}\dataTrainTestCode

\pgfplotstableread{
Year  CVPR  ICCV  ICLR  ICML  NeurIPS  ALL
2021  25.2  28.2  19.3  10.8  14.1     19.5
2022  31.4  nan   21.6  12.3  15.3     20.5
2023  28.9  30.1  21.7  14.2  17.2     22.3
2024  29.3  nan   21.9  15.7  21.0     21.9
2025  26.0  25.4  22.6  15.9  16.7     20.6
}\dataWeights

\pgfplotstableread{
Year  CVPR  ICCV  ICLR  ICML  NeurIPS  ALL
2021  15.1  17.2  09.2  05.3  09.7     11.7
2022  20.5  nan   14.7  09.4  11.7     14.4
2023  24.6  25.1  15.4  11.3  14.3     18.2
2024  25.5  nan   17.5  14.3  19.0     19.2
2025  26.1  24.2  19.4  14.5  14.3     18.7
}\dataAckRepos


\nextgroupplot[title={(a) Proportion of accepted papers with train or test code}]
  \addplot+ table[x=Year,y=CVPR]{\dataTrainTestCode};
  \addplot+ table[x=Year,y=ICCV]{\dataTrainTestCode};
  \addplot+ table[x=Year,y=ICLR]{\dataTrainTestCode};
  \addplot+ table[x=Year,y=ICML]{\dataTrainTestCode};
  \addplot+ table[x=Year,y=NeurIPS]{\dataTrainTestCode};
  \addplot+ table[x=Year,y=ALL]{\dataTrainTestCode};

\nextgroupplot[title={(b) Proportion of accepted papers with available model weights}]
  \addplot+ table[x=Year,y=CVPR]{\dataWeights};
  \addplot+ table[x=Year,y=ICCV]{\dataWeights};
  \addplot+ table[x=Year,y=ICLR]{\dataWeights};
  \addplot+ table[x=Year,y=ICML]{\dataWeights};
  \addplot+ table[x=Year,y=NeurIPS]{\dataWeights};
  \addplot+ table[x=Year,y=ALL]{\dataWeights};

\nextgroupplot[title={(c) Proportion of accepted papers acknowledging other repositories}, xlabel={Year}]
  \addplot+ table[x=Year,y=CVPR]{\dataAckRepos};
  \addplot+ table[x=Year,y=ICCV]{\dataAckRepos};
  \addplot+ table[x=Year,y=ICLR]{\dataAckRepos};
  \addplot+ table[x=Year,y=ICML]{\dataAckRepos};
  \addplot+ table[x=Year,y=NeurIPS]{\dataAckRepos};
  \addplot+ table[x=Year,y=ALL]{\dataAckRepos};

  \legend{CVPR,ICCV,ICLR,ICML,NeurIPS,All}

\end{groupplot}
\end{tikzpicture}

}
\caption{Proportion of accepted papers at CVPR, ICCV, ICLR, ICML and NeurIPS 2021–2025 that publicly (a)~share train or test code [reducing], (b) provide pretrained model weights [reducing], or (c)~acknowledge previous repositories [growing].}
\label{fig:ratio_code}
\vspace{-20pt}
\end{wrapfigure}
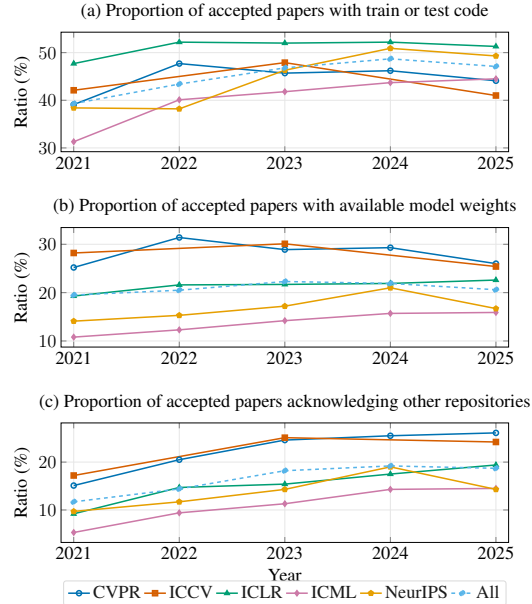

\paragraph{Common reasons why software is unavailable.}

In our experience, there are three main reasons explaining (rather than justifying) why some code or data is not available.
\begin{itemize}[
    leftmargin=*,
    labelsep=0.35em,
    itemsep=2pt,
    topsep=-2pt,
    parsep=0pt,
    partopsep=0pt
]
    \item \emph{No time to polish}
    is not a legit argument. If the software is not clean, chances are that the results are not either, hampering reliability. 

    \item \emph{The developer left} is not a valid argument either. Software polishing should be an integral part of the developer's work, including if s/he is an intern bound to leave after a few months.

    \item \textit{The company does not want to disclose software}, for legal or security reasons, or to make it more difficult for competitors to reproduce the work. The scientific implication is debatable: either the paper contains enough information to be reimplemented and the company only gains a little time over competitors, or it does not and the paper has little scientific value as it is not reproducible. A usually accepted compromise is to only provide learned parameters and inference code, possibly with a binding license.
    In contrast, academic papers are expected to give full access to artifacts for the sake of open science \cite{d2025lessons}.
\end{itemize}

\subsection{Code availability in ML/CV conferences}
\label{sec:studycode}

\begin{figure}
\centering
\resizebox{\linewidth}{!}{

\begin{tikzpicture}

\pgfplotstableread[col sep=space]{
Year  T1   T2   T3   T4   T5   T6   T7   T8    T9   T10
2021  17.5  8.4  7.2  8.6  1.4 13.2 21.1 16.3  1.2  1.7
2022  13.2  5.8  6.6 11.0  2.0 14.4 22.8 18.2  1.3  1.4
2023  17.7  5.7  9.0 12.4  5.2 11.0 20.2 13.4  0.8  1.7
2024  12.8  4.2  8.7 21.6  8.7  9.8 16.6 12.8  0.7  1.5
2025  13.0  4.4 13.2 24.9 10.6  7.5 13.4  9.0  0.5  1.3
}\topicshare

\pgfplotstableread[col sep=space]{
Year  T1   T2   T3   T4   T5   T6   T7   T8    T9   T10
2021  36.0  47.4  39.3  47.3  48.2  33.6  41.4  27.0  44.7  39.1
2022  41.5  50.9  52.2  48.2  47.1  32.3  44.7  32.6  50.5  60.4
2023  44.0  52.1  49.0  50.1  52.3  37.8  45.8  33.2  57.0  56.0
2024  47.0  48.3  47.2  47.8  51.5  41.7  44.5  36.2  49.4  48.8
2025  37.4  41.2  41.2  43.7  42.5  42.6  41.8  33.7  39.8  53.0
}\topicsharecode

\begin{groupplot}[
  group style={
    group size=2 by 1,        
    horizontal sep=1.9cm,
    ylabels at=edge left,
    yticklabels at=edge left,
  },
  width=\textwidth,
  height=0.4\textwidth,
  symbolic x coords={2021,2022,2023,2024,2025},
  xtick=data,
  xlabel={Year},
  ymin=0, ymax=65, 
  scaled y ticks=false,
  yticklabel style={/pgf/number format/fixed, /pgf/number format/precision=0},
  grid=both,
  grid style={black!10},
  axis line style={black!55},
  tick style={black!55},
  every axis plot/.append style={thick, mark size=1.6pt},
  cycle list name=exotic,
    tick label style={font=\Large},
label style={font=\Large},        
title style={font=\Large},
  legend style={
    at={(-0.13,-0.32)},
    anchor=north,
    draw=black!15,
    fill=white,
    fill opacity=0.9,
    rounded corners=2pt,
    font=\Large,
    cells={align=left},
  },
  legend columns=4, 
  clip=false,
  title={\TODO{Topic evolution across all conferences (2021--2025)}},
  cycle list={
      {color=topic1,  very thick, mark=o},
      {color=topic2,  very thick, mark=square*},
      {color=topic3,  very thick, mark=triangle*},
      {color=topic4,  very thick, mark=diamond*},
      {color=topic5,  very thick, mark=pentagon*},
      {color=topic6,  very thick, mark=*},
      {color=topic7,  very thick, mark=otimes*},
      {color=topic8,  very thick, mark=x},
      {color=topic9,  very thick, mark=star},
      {color=topic10, very thick, mark=+},
  },
  enlarge x limits=0.03,
]

\nextgroupplot[title={(a) Evolution of topics in years},
ymin=0, ymax=25, ylabel={Topic share (\%)}, xlabel={},]
\addplot+ table[x=Year,y=T1]{\topicshare};  
\addplot+ table[x=Year,y=T2]{\topicshare};  
\addplot+ table[x=Year,y=T3]{\topicshare};  
\addplot+ table[x=Year,y=T4]{\topicshare};  
\addplot+ table[x=Year,y=T5]{\topicshare};  
\addplot+ table[x=Year,y=T6]{\topicshare};  
\addplot+ table[x=Year,y=T7]{\topicshare};  
\addplot+ table[x=Year,y=T8]{\topicshare}; 
\addplot+ table[x=Year,y=T9]{\topicshare}; 
\addplot+ table[x=Year,y=T10]{\topicshare}; 

\nextgroupplot[title={(b) Code availability ratio in each topic},
ymin=25, ymax=65, ylabel={Code availability (\%)},]
\addplot+ table[x=Year,y=T1]{\topicsharecode};  
\addlegendentry{3D Vision and Neur. Render.}
\addplot+ table[x=Year,y=T2]{\topicsharecode};  
\addlegendentry{Image and Video Synt.}
\addplot+ table[x=Year,y=T3]{\topicsharecode};  \addlegendentry{Multimodal Learning}
\addplot+ table[x=Year,y=T4]{\topicsharecode};  
\addlegendentry{LLMs and Found. Models}
\addplot+ table[x=Year,y=T5]{\topicsharecode};  
\addlegendentry{Gen. AI and Diff. Models}
\addplot+ table[x=Year,y=T6]{\topicsharecode};  \addlegendentry{Reinforcement Learning}
\addplot+ table[x=Year,y=T7]{\topicsharecode};  \addlegendentry{Trust., Fair and Robust ML}
\addplot+ table[x=Year,y=T8]{\topicsharecode}; 
\addlegendentry{Optim. and Theory of Learn.}

\addlegendimage{empty legend}\addlegendentry{} 

\addplot+ table[x=Year,y=T9]{\topicsharecode}; 
\addlegendentry{Scal., Syst. \& Distrib. ML}

\addplot+ table[x=Year,y=T10]{\topicsharecode}; 
\addlegendentry{Application-Driven ML}
\addlegendimage{empty legend}\addlegendentry{} 


\end{groupplot}
\end{tikzpicture}

}
\vspace*{-15pt}
\caption{Proportion of papers per topic 
in accepted papers to CVPR, ICCV, ICLR, ICML and NeurIPS 2021–2025 (left), and code availability in each topic (right). Except for a few topics, the general trend shows a proportional decrease in open-sourcing official codes.}
\label{fig:topics}
\vspace{-2mm}
\end{figure}
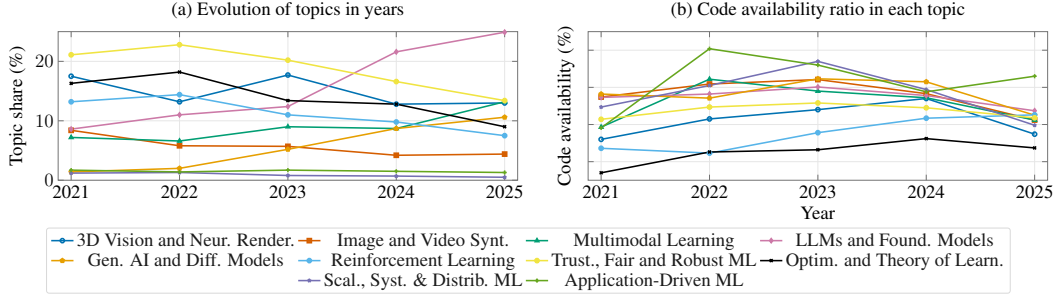

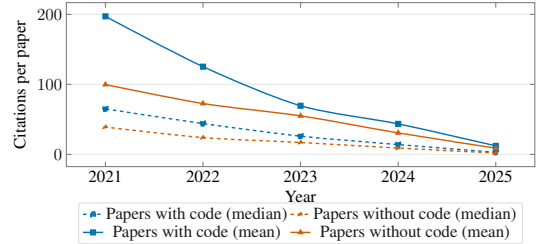
\begin{wrapfigure}{r}{0.5\textwidth}
\vspace{-10pt}
\resizebox{\linewidth}{!}{

\begin{tikzpicture}

\pgfplotstableread[col sep=space]{
Year  All  WithCode  WithoutCode
2025  3   3        2
2024  11   14       9
2023  21   26       17
2022  31   44       24
2021  48   65       39
}\citdatamedian

\pgfplotstableread[col sep=space]{
Year  All  WithCode  WithoutCode
2025  10.42839582052858  12.172814612738703      8.874709486583562
2024  36.95112296704242  43.55530035335689       30.680422747861098
2023  61.75691948238677  69.44395393474089       54.989523487664755
2022  95.50042432814710  125.27074520013016      72.61220915686765
2021  138.0243169041705  197.16251246261217      99.67370689655172
}\citdatamean

\begin{axis}[
  scale only axis,
  width=\textwidth,
  height=0.35\textwidth,
  symbolic x coords={2021,2022,2023,2024,2025},
  xtick=data,
  xlabel={Year},
  ylabel={Citations per paper},
  ymajorgrids,
  grid style={black!10},
  axis line style={black!55},
  tick style={black!55},
  scaled y ticks=false,
  yticklabel style={/pgf/number format/fixed, /pgf/number format/precision=0},
  smooth,
  every axis plot/.append style={very thick, mark size=2pt},
  cycle list={
    {color=cCVPR,  dashed, mark=square*},
    {color=cICCV,   dashed, mark=triangle*},
    {color=cCVPR,  mark=square*},
    {color=cICCV,   mark=triangle*},
  },
      tick label style={font=\Large},
label style={font=\Large},        
title style={font=\Large},
  legend style={
    at={(0.5,-0.22)},
    anchor=north,
    draw=black!15,
    fill=white,
    fill opacity=0.9,
    rounded corners=2pt,
    font=\Large,
    cells={align=left},
  },
  legend columns=2,
  clip=false,
]


\addplot+ table[x=Year,y=WithCode]{\citdatamedian};
\addlegendentry{Papers with code (median)}

\addplot+ table[x=Year,y=WithoutCode]{\citdatamedian};
\addlegendentry{Papers without code (median)}


\addplot+ table[x=Year,y=WithCode]{\citdatamean};
\addlegendentry{Papers with code (mean)}

\addplot+ table[x=Year,y=WithoutCode]{\citdatamean};
\addlegendentry{Papers without code (mean)}

\end{axis}
\end{tikzpicture}
}
\caption{Mean \& median citation counts for publications at CVPR, ICCV, ICLR, ICML and NeurIPS (2021–2025), grouped by code availability.
}
\label{fig:citations}
\vspace{-5mm}
\end{wrapfigure}

To quantify code availability, we conducted a statistical study on the accepted papers of recent (2021-2025) top-tier conferences in ML/CV, including CVPR, ICCV, ICLR, ICML, and NeurIPS. The main collected data are represented in \cref{fig:ratio_code,fig:topics,fig:citations}. More details are available in \appref{app:study_details}.
\paragraph{Code sharing has recently started to decrease.}

After several years of noticeable increases in the proportion of accepted papers with (train or test) code publicly available, a marked decline is observed in 2025 for all conferences (\cref{fig:ratio_code}(a)). The deterioration even dates back to 2023 for CVPR and ICLR. Besides, this drop occurs while the proportion of papers with topics calling for empirical validation is significantly rising, as opposed, for instance, to more theoretical topics (\cref{fig:topics}). A similar decline is visible for accepted papers with model weights, although it is not as marked (\cref{fig:ratio_code}(b)).

Additionally, the 20\% most acknowledged repositories in our study (\cref{fig:ratio_code}(c)) account for 81.5\% of all acknowledgements, following the Pareto principle. Looking at the most acknowledged repos, it appears that a significant part of the recent progress in empirical machine learning originates from foundation models that make weights available, although not always training code and data.

\paragraph{Papers with code are cited twice as often.}

We also collected citation counts from \citet{semantic_scholar} for all the accepted papers and calculated the mean and median citation values for the papers with and without public code (\cref{fig:citations}). We observe that papers with official public code implementations have more than twice as many citations on average as those without any code. While we understand that the community is not ready to require code availability for publications, and that our analysis is correlational rather than causal, this observation should nevertheless encourage authors to share their code. 

\paragraph{Code reuse stagnates but still slightly boosts research.}

Finally, we counted the number of repositories of accepted papers that acknowledge other repositories (\cref{fig:ratio_code}(c)). We consider that it typically corresponds to implementations that borrow or modify some existing code to build their own. We observe that the number of acknowledging repositories has not changed significantly over the past three years, although there are differences between conferences that balance each other out. Relative to the drop in accepted papers with code (\cref{fig:ratio_code}(a)), this stagnation means that public implementations have nevertheless slightly increased their capacity to boost research.

\section{Code availability does not ensure paper reproducibility}
\label{sec:barriers}

Code is thus not always available. But even if so, it does not ensure reproducibility.
In this section, we study inconsistencies between code and paper (\cref{sec:issues_paper_vs_code}), and possible execution issues (\cref{sec:issues_running_code}). 

\subsection{Discrepancies between paper and code}
\label{sec:issues_paper_vs_code}

Getting a hold on the code is not just about obtaining the same numbers as those in the paper. It is also about making sure the code does what the paper says. It is the paper that conveys the ideas and that provides a rationale to evaluate them; the code serves as support, and it must be a faithful one.

Yet, due to page limitations, some implementation details are commonly found only in the code, rather than in the paper. As newly proposed architectures are increasingly complex, reimplementing a paper has thus become a real challenge. Besides, it is not uncommon to find differences between the paper description and what the code actually does. It impairs reproducibility, making it difficult for reimplementations to match the original results.
It misleads the reader and thus also impairs progress.

Conversely, some processing can be missing from the code that is made public, although related results are reported in the paper. It can be the case, e.g., of test-time augmentations and ensembling, which are sometimes used to boost a metric to reach the SOTA on a specific benchmark, while being barely mentioned in the paper. In our experience, e.g., for semantic segmentation, such precious processing tricks can remain undisclosed and thus unreproducible. It boosts the performance on a benchmark with server and hidden ground truth, while the basic version of the method, showcased in the paper on another dataset with public ground truth, is on the contrary fully reproducible.

There may also be methodological mistakes in the code, which are not visible in the paper but that could invalidate the reported experiment. For instance, a model can be trained by introducing more data or information than intended and indicated. 
Concrete bad practices include to pretrain unsupervisedly on test data, 
or to train on test data using a semi-supervised approach.

\subsection{Issues when running code}
\label{sec:issues_running_code}

Even if the code is sensible and consistent with the paper, some issues may occur at execution time. 

Execution is often stochastic: during training (model initialization, batch randomness, asynchronous operations...) but also at test time (e.g., GenAI). Yet, not all authors present metrics averaged over several runs, with an indication of variance. This may lead to reproduction differences, which are however often accepted if the variations stay moderate, depending on the community, task and dataset.

There are also bad practices related to training monitoring. A classical one consists of peeking at the test set, quantitatively if the ground truth is known, or qualitatively otherwise. Training can then be stopped early to prevent overfitting, as often done, e.g., on target sets for supposedly unsupervised domain adaptation, instead of using validators \cite{musgrave2022benchmarking}. Another bad practice is cherry-picking a best checkpoint among several training attempts, possibly explaining results that nobody succeeds in reproducing. Besides, when only model parameters are given, it is hardly possible to know what learning scheme and actual data were used for training. As for inference, difficult samples can be excluded, as we already witnessed. Test-time augmentation or ensembling can be used but unreported.

Last, a practical issue, besides installation problems, is that the (train or test) code might be too long to run, or require specific hardware (e.g., lots of high-end GPUs).

\section{Verifiable indicators of reproducibility}
\label{sec:checking}

To err is human. But in the era of (M)LLMs, ensuring that papers present genuine experiment outputs rather than erroneous or fake results is more critical than ever. As code, even if accessible, is in any case difficult to assess, we propose instead to focus on the verification of easily-available indicators.

The problem is that any execution that can only be performed by the authors is a source of vulnerability regarding reproducibility and verifiability. The most reliable way to enable the validation of an experimental result remains to open-source the training and inference code, the data, and the trained models. However, we recognize that open-sourcing is sometimes impossible due to privacy constraints, licensing, commercial restrictions, or security concerns. 
To maintain a healthy, trustworthy research environment and prevent bad research practices, alternative verification methods are required.

In this section, we describe two verifications that can contribute to improving confidence in reported experiments: checking experiment logs (\cref{sec:log_checking}) and checking metric evaluations (\cref{sec:metric_checking}).
The complete associated workflow is presented in \cref{sec:policy}. We also discuss code release (\cref{sec:code_release}).

\subsection{Verification of experiment logs}
\label{sec:log_checking}

\paragraph{Log verification material.}
Most, if not all experimental environments used in the ML community support the generation of logs for monitoring the experiments, including training and inference. Such environments include TensorBoard~\cite{tensorboard}, MLflow~\cite{zaharia2018mlflow}, and Weights \& Biases (W\&B)~\cite{wandb}. 
Logs to provide for verification purpose concern each experiment related to empirical claims in the paper, in particular state-of-the art (SOTA) claims, including via tables and graphs. Ablation experiment logs are optional. 
Provided log data should include a list of basic information on the experiment, as described in \appref{app:examplelog}, and     learning curves of trained models (examplified also in \appref{app:examplelog}).

\paragraph{Providing log verification material.}

Experiment logs are to be provided by the authors at paper submission time, in the supplementary material. To ease verification, authors should also include any relevant plots, exported from the logs.
Contrary to the code, experiment logs hardly raise any issues regarding privacy, licensing or security. Their disclosure should thus be easily accepted. Since logging useful training and inference metrics with free tools is already common practice for model analysis, the additional burden on authors is minimal, e.g., similar to filling the NeurIPS checklist.

\paragraph{Checking log verification material.}

The consistency of the experiment logs with the paper is checked by the reviewers during the review period. If the conference features a rebuttal, the reviewer may ask the authors for clarification or additional log information. The report on log consistency is part of the review. (More details are in \cref{sec:policy}.)
While CVPR 2026 has introduced optional W\&B log uploads, the focus is on compute consumption. Our proposal is to use it for experiment verification.

\paragraph{Goal and limitations of log verification.}

Log-based verification ensures that reported numbers are backed by real training and evaluation traces. But it does not fully prevent data leakage, dataset manipulation, or ``lookup cheating'', where predictions are recovered using sample identifiers.

\subsection{Verification of metric computation}
\label{sec:metric_checking}

\paragraph{Metric verification material.}

Most experiments in ML try to quantify the quality of predictions, including stochastic generations. It is performed by running some code that computes metrics over result files, typically by measuring a form of distance to some ground truth, including distribution models. To make sure no error is introduced when computing these metrics and when reporting them, each experiment related to a claim in the paper should come with \textit{results files}, \textit{evaluation code}, and possibly ground truth information. (See details in \appref{app:metricverif}.) No license agreement should be needed to access proprietary data because it could break the anonymity of the authors or of the reviewer.

\paragraph{Providing metric verification material.}

The metric verification material is to be provided by authors at paper submission time, in the supplementary material or via links to anonymized repositories. Even though there may be cases where the output of a model is sensitive and cannot be disclosed, results most often concern publicly available datasets and do not have disclosure issues. Reviewers should be able to inspect the evaluation material and rerun it easily, avoiding installation issues. We therefore recommend that authors actually provide an anonymized notebook, as available in free shared execution environment such as Google Colab, Kaggle Notebooks, Paperspace Gradient, or AWS SageMaker Studio Lab. We provide two examples of such notebooks~\footnote{https://github.com/giddyyupp/position-enforce-verifiability}: a short script using only public libraries and a longer script containing a custom metric implementation. Preparing such metric evaluation code and data for review introduces only minimal overhead for authors, as it mostly involves reusing code they already have written for their own evaluation purpose.

\setlength{\columnsep}{12pt}
\begin{wraptable}[17]{r}{0.5\textwidth}
    \centering
    \vspace{-12pt}
    \caption{Provided indicators of reproducibility.} 
    \label{tab:reprod_indic}
    \resizebox{0.5\textwidth}{!}{
    \begin{tabular}{l|c@{~~}l}
    \toprule
    Indicator & \multicolumn{2}{l}{Status or rating} \\
    \midrule
    \multirow{4}{*}{Experiment logs}
        & \cmark & consistent with paper\\
        & \qmark & inconclusive wrt paper \\
        & \xmark & inconsistent with paper \\
        & \nmark & unavailable (default) \\
    \midrule
    \multirow{4}{*}{Checked metrics}
        & \cmark & similar enough \\
        & \qmark & questionably different \\
        & \xmark & definitely different \\
        & \nmark & unavailable (default) \\
    \midrule
    \multirow{3}{*}{Software}
        & \cmark & link to a repository \\
        & \xmark & proprietary \\
        & \nmark & unmaintained (default) \\
    \bottomrule
    \end{tabular}}
\end{wraptable}

\paragraph{Checking metric verification material.}

To reduce the overall workload and depending on the venue policy, one or several of the reviewers are to be appointed as \emph{metric evaluators} for the same paper. The responsibility of a metric evaluator is to scrutinize the metric evaluation code, to run it on provided result files, to check metric consistency with the paper, and to write a comment about it as part of the paper review. 
We conducted a small user study with five experienced reviewers to measure the average time required to apply the verification instructions.
The short script required approximately 5 minutes on average, while the longer script required between 45 minutes and 1 hour. Overall, this workload is substantially lighter than reviewing a full paper and providing detailed feedback. We consider this an acceptable cost to gain trust in empirical results. 

\paragraph{Goal and limitations of metric verification.}

This verification ensures that the metrics reported in the paper are identical (up to acceptable stochastic variations) to those obtained by an independent evaluator. While the result files may be forged, this verification, and the care required for authors to prepare it, should help reduce mistakes made when quantifying a result and reporting it.

\subsection{Verification of code release}
\label{sec:code_release}
After acceptance, authors may provide code, typically by inserting in their paper a link to a repository. There is currently no verification that such links actually point to sensible code. Still, reviewers sometimes treat code promises positively, even though such commitments can be fragile. (It is easy to find GitHub repositories of papers from our five target conferences, that contain the (in)famous ``code coming soon'' and that are at least one year old, if not much more.)

To promote timely code release, we propose that, by the camera-ready deadline, the authors provide a software link if they wish, and that, by the time the conference opens, an assigned reviewer checks the repository to verify it does contain the expected software. As the goal is not to scrutinize the software but just to check that it exists, the task is very lightweight. It can actually by largely automated, as we did in \cref{sec:studycode}, with a possible human control for failing repos.
This official software link, if any and if confirmed by the reviewer, would appear on the proceedings web site, as done, e.g., for ICML in \citet{pmlr}. Although, software completeness will not be checked, we believe it would nonetheless push authors not to delay the code release, and operate as an incentive to give value to their paper.

\begin{figure*}[t]
    \centering
    \resizebox{0.94\textwidth}{!}{
        \input{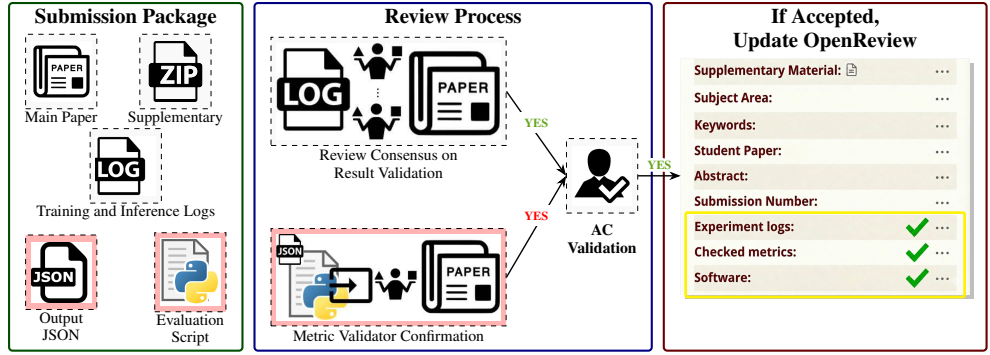}
    }
    \caption{Overview of the proposed verification workflow. At submission time, \textit{training and inference logs} are attached to the submission package and verified by one or several reviewers (Level~1). The metric verification step (Level~2), indicated with red background, requires the submission of JSON file(s) that contain model predictions on target dataset(s), along with an evaluation script. The assigned metric validator runs the evaluation script and compares with the numbers in the main paper. Finally, the AC updates the corresponding statuses (\cref{tab:reprod_indic}) in the submission site.}
    \label{fig:proposal}
\end{figure*}

\section{Call to action: a reinforced review process}
\label{sec:policy}

To improve trust in accepted papers, we propose a verification workflow (\cref{fig:proposal}) that is realistic, implementable immediately, and low-overhead for conferences, authors and reviewers.

\textbf{When a paper is submitted:}
\newcounter{policy}
\begin{itemize}[topsep=0pt,parsep=1pt,itemsep=1pt,leftmargin=2.5em]
\stepcounter{policy}\item[P\arabic{policy}.]
    \textbf{The authors must} make available the experimental material (logs, results and evaluation code, \cref{sec:checking}) of their main experiments, i.e., related to claimed contributions. It can be given via anonymized links or provided in supplementary material on the submission web site.
    
\end{itemize}

\textbf{When a paper is reviewed and discussed}:
\begin{itemize}[topsep=0pt,parsep=1pt,itemsep=2pt,leftmargin=2.5em]

\stepcounter{policy}\item[P\arabic{policy}.]
    \textbf{The reviewers must} comment on the consistency of the provided \textit{experimental material} w.r.t.\ the empirical results presented in the submitted paper. The reviewers must also give a formal rating on this experimental consistency, as defined in \cref{tab:reprod_indic}. This is an integral part of the review, which contributes to the final recommendation. 
    To lighten the overall workload, a venue policy can be to assign a single \textit{experiment reviewer} per paper.

    \stepcounter{policy}\item[P\arabic{policy}.]
    \textbf{The area chairs must} summarize the assessment of the experimental consistency and provide a rationale for a final consistency status. It is an integral part of the meta-review.
    \stepcounter{policy}\item[P\arabic{policy}.]
    \textbf{The reviewers and area chairs may} value more, when reviewing a paper, a comparison with another paper that has been reported to have experimental consistency.
    
\stepcounter{policy}\item[P\arabic{policy}.]
    \textbf{The program chairs must} give each submitted paper an experimental consistency status, based on the recommendations of the area chairs. It is an integral part of the final decision.

\end{itemize}

\textbf{When a paper is accepted:}
\begin{itemize}[topsep=0pt,parsep=1pt,itemsep=2pt,leftmargin=2.5em]

\stepcounter{policy}\item[P\arabic{policy}.]
    \textbf{The authors must} provide, before the camera-ready deadline, a status as defined in \cref{tab:reprod_indic} regarding the software (code and/or data, including model weights) that supports the empirical results in their paper. This software status will be visible on the proceedings web site, as well as the experimental consistency rating.
    If a link to a repository is provided, the software must be readily available, although possibly after signing a license agreement. 
\end{itemize}

\textbf{When a paper is published:}
\begin{itemize}[topsep=0pt,parsep=1pt,itemsep=1pt,leftmargin=2.5em]
    \stepcounter{policy}\item[P\arabic{policy}.]
    \textbf{The program chairs must} make visible on the submission 
    or proceedings web sites:
    the experimental consistency rating, the software availability status,  the experimental material. 
\end{itemize}

\textbf{When a paper is written:}
\begin{itemize}[topsep=0pt,parsep=1pt,itemsep=1pt,leftmargin=2.5em]
    \stepcounter{policy}\item[P\arabic{policy}.]
    \textbf{The authors of a new paper may} structure their arguments, tables and graphs to highlight results and comparisons to papers that are stamped as experimentally consistent or that provide software reproducing the experiments, separating them from unlabeled other papers.
    (Papers published before this policy have an `unavailable' experimental consistency status.)
\end{itemize}

We note that the experimental consistency is limited to submitted logs and/or metric-evaluation material with the reported results. This status does not certify full reproducibility, the correctness of the training or data pipeline, the authenticity of the outputs, or overall reliability.

\boldparagraph{Exemption}
Not providing verification material is possible, but a justification must then be provided, which reviewers and area chairs will assess. It may concern theoretical papers (which may however have motivating experiments), experiments with hidden-test benchmark servers, experiments on private data (which reviewer may however not value as highly), or excessively large output files. In contrast, copyright (vs private) data is not an excuse for not providing evaluation material as reviewers already commit to treat submissions as confidential.

\boldparagraph{Implementation} This policy can easily be implemented in OpenReview:
(i)~It already hosts many major ML and CV conferences; (ii)~More conferences continue to adopt it due to its flexibility and open-source nature; (iii)~It is actively supported by the research community.
Moreover, OpenReview is easily and commonly tailored according to the specific requirements of different venues. Final statuses (\cref{tab:reprod_indic}) and code links may also appear on the web sites of NeurIPS, PMLR and The CVF.

This policy can be implemented incrementally, first adopting Level~1 (log verification), then when Level~1 is established, introducing Level~2 (metric checking) towards a stronger validation.

\section{Alternative views}
\label{sec:alternativeviews}

\textbf{Imposing code submission.} 
Some conferences, such as VLDB, require code to be submitted \cite{vldbendowment}. So does ASIACCS, although there was no code evaluation process this year \cite{asiaccs2026} and a valid reason for not doing so could be provided. Journals such as IPOL \cite{ipol} also tightly couple papers and code.

\textbf{Incentivizing code submission.}
ICPR prefers an incentive to promote code submission, with a Reproducible Research in Pattern Recognition (RRPR) Badge \cite{icprbadge}, which has been introduced since 2016. The evaluation criteria are at the discretion of specially-designated RRPR reviewers, and Reproducibility Chairs add their own meta-review to the reproducibility reviews. This review is performed only on accepted papers, thus having no impact on acceptance decisions. 

\textbf{Encouraging code submission.}
NeurIPS, ICLR, CVPR and ICCV encourage code submission, which reviewers are ``welcome'' to read, but ``not required'' to. ICML allows code submission too, but is somewhat ambiguous: while ``reproducibility of results and easy availability of code will be taken into account in the decision-making process'', ``it is entirely up to the reviewers to decide whether they wish to consult any of the appendices in the submitted paper or the supplementary material''. However, the current practice is not full-fledged code evaluation but code browsing for missing details\rlap.

\textbf{The code virtue hypothesis.} A previous reader of this position paper thinks that authors ``should provide code, and that should be its own incentive'' because ``it better advances science'', as also emphasized in \cite{donoho2024datascience}. However, while we note that publications with code have twice as many citations (\cref{sec:studycode}), we also observe a stagnation in code availability, between 40\% and 50\% of published papers, depending on venues (\cref{sec:studycode}). Besides, code success as an a posteriori assessment does not prevent some ``toxic'' SOTA but codeless  papers from hindering the publication of new methods which would not reach this SOTA. In contrast, we propose an a priori verification.

\textbf{No code, no cite.} Still more radical than our proposal, we know of colleagues who have decided in their own publications to never compare to any paper without code. Our position is not quite so clear-cut: we do not ban codeless papers, but prefer to highlight methods coming with code.

\textbf{De-emphasizing reproducibility.} Contrary to the above, \citet{drummond2018reproducible} argues that reproducibility is not essential to science, that requiring code to be submitted is unnecessary and would even reduce paper acceptance to narrow technical criteria, and that misconducts actually have little impact.

\textbf{De-emphasizing experiments.} Not exactly contradictory, result-blind peer review focuses on the ideas presented in the paper regardless of any empirical evidence, with a submitted paper deprived from results and conclusion \cite{rosenthal1966}. The full paper is to be provided in a second reviewing stage \cite{mahoney1977publiprejudices}. 

\textbf{Code review burden.} Reviewing code, that took weeks or months to be written, is a heavy burden for reviewers, which probably explains why the RRPR badge is awarded via a specific committee. One might thus consider that the workload is too high a price to pay for the added guarantee.
Our proposal however only concerns the evaluation code, which is usually small, if not already in libraries.

\section{Perspectives}
\label{sec:perspectives}

Our proposal is not bullet-proof, but it should still make the lives of fraudsters a bit harder. It would also make their infractions more explicit and less forgivable if they were discovered.

To go beyond this log and metric checking would require code available at review time. However, the decision for a venue to ignore possible reasons not to disclose code and impose code submission seems to be hard to make. 

We believe that checking the consistency of code with respect to a paper is actually a long shot, if only 
because of the significant reviewer workload and compute it would require.

Still, Paper2Code \cite{seo2025paper2code}, which proposes a first approach to automatically generate code from papers, interestingly also proposes an LLM-based method to evaluating how well a code repository reflects the contents of a paper, with a level of performance close to that of a human. This verification task is indeed simpler than code generation, and we may expect in the near future automated systems to provide reliable reports on the consistency between code and paper. This should alleviate the load of \emph{reproducibility reviewers}, who could concentrate on possible issues mentioned in generated consistency reports. Security would, however, have to be guaranteed so that proprietary code cannot leak via LLM requests. It would be up to program chairs to provide reviewers with reports produced by a secure LLM, keeping a human in the loop for interpretation.

Another perspective is for the authors to upload code, models, and data to an independent third-party server, secured to allow uploads of proprietary software. The server would run the code and automatically generate a report, while an LLM-based auditor would inspect the code for suspicious behavior, such as hard-coded outputs, lookup tables, or other shortcuts that could artificially reproduce the claimed numbers. Unfortunately, this framework is unlikely to be feasible in the near future because hosting, securing, and maintaining such a platform is expensive and operationally complex.

\section{Conclusion}
\label{sec:conclusion}

Our study shows that, after a period of marked improvement, the ability to reproduce empirical results of publications in top-tier ML and CV conferences has recently deteriorated. This observation comes at a time when the capacity of (M)LLMs is increasing dramatically, changing the way code is developed and papers are written and reviewed, while casting doubt on the reliability of publications.

Although code and data availability remains the key to evaluating empirical results in computer science publications, many authors are still reluctant to make their software accessible. Considering that providing enough details to make a paper really reproducible and assessable by reviewers is not very far from providing actual code should defeat arguments against code availability. Still, the current consensus of the scientific community is to accept papers coming without supporting software, hence papers that could be hard to reproduce and thus to fully evaluate in the first place.

In this context, while still promoting code availability for reproducibility and research acceleration, we approach the problem from the \textit{verifiability} perspective, which is much lighter than actual \textit{reproducibility}, although it remains efficient. We propose a new policy to improve the ability to evaluate empirical results, based on the verification of experiment logs and metric computation, without the need for actual code, which does not guarantee replication anyway. And this policy seamlessly integrates into the existing reviewing and publication process.

There is no free lunch tough; our proposal requires some additional effort from authors and reviewers. We however quantify it and argue it is minimal. We believe this light extra workload is worth it, because it can significantly improve the quality and reproducibility of published papers.

This is an initial proposal for establishing a new practice for evaluating empirical results and citing them. We hope the community can discuss it, improve it, and start implementing it in upcoming conferences, at least as an experiment.

\section{Acknowledgment}
We acknowledge the EuroHPC Joint Undertaking for awarding the project ID EHPC-REG-2024R02-234 access to Karolina, Czech Republic.

\bibliography{reprod}
\bibliographystyle{unsrtnat}

\clearpage
\newpage
\appendix

\section*{Appendix}

In this appendix, we provide:
\begin{itemize}[nosep]
    \item details on our study regarding code availability and citations (\appref{app:study_details}),
    \item details and examples regarding the verification framework (\appref{sec:detailverifframework}),
    \item an additional discussion regarding code, before and after acceptance (\appref{app:code}).
\end{itemize}
We also wish to recall that our proposal is a starting point, not a set-in-stone procedure. Thanks to this position paper, we hope the community will be able to debate the question, improve the proposition into a clear Instruction Guide and FAQ, and implement it at an upcoming event (e.g., workshop) on a trial basis before it can be applied more broadly and at a larger scale. To gain credibility and be in a position to convince the organizers of a venue to support a pilot study, we also believe our proposal will benefit from a recognition as a position paper in a major conference like NeurIPS.

\section{Details of the study on code availability}
\label{app:study_details}

In this section, we explain our study on code availability (\cref{sec:studycode}) in detail. Our focus is on the top-tier Computer Vision (CV) and Machine Learning (ML) conferences, i.e., CVPR, ICCV, ICLR, ICML and NeurIPS. We also narrow our scope to the last 5 years, \ie, from 2021 to 2025 inclusive. 

We first downloaded the PDFs (main paper and supplementary) of all the accepted papers from their official website (\href{https://www.thecvf.com/}{thecvf.com}, \href{https://iclr.cc/}{iclr.cc}, \href{https://icml.cc/}{icml.cc}, \href{https://neurips.cc/}{neurips.cc}).  
Then, we analyzed their contents to detect possible code URLs. We extracted all the URL in the paper using a generic regular expression and fed them, along with the context surrounding them, to an LLM to determine which URL is more likely to be the implementation of the method described in the paper. We present the prompt used during this process in Prompt~\ref{prompt:url_prompt}. 

In the next stage, we tried to clone into a local directory the code URL of each paper, if available. It is common practice in CV/ML papers to provide a project page containing a link to the code repository along with other useful information, such as visual results. To overcome the cloning failures in such cases, we added a resolver function if the detected URL is not a ``Git" repository but a project page. This resolver looks for any repository link on the project page.  

For each cloned repository, we again prompted the LLM using the downloaded files and the code URL to tell:

\begin{itemize}[nosep]
    \item If the repository is training based,
    \item If specific training is required,
    \item If the training code is available,
    \item If the inference or test code is available,
    \item If the training instructions are available,
    \item If the model weights are available,
    \item If there are open issues regarding the installation or reproducibility of results in the paper,
    \item If other repositories are acknowledged, suggesting that external code may have been borrowed and adapted.
\end{itemize}

In Prompt~\ref{prompt:audit_prompt}, we present the prompt used to obtain this assessment of the code repositories.

\begin{table}
\caption{Detailed statistics for CVPR.}
\label{tab:cvpr_detailed_stats}
\rowcolors{4}{gray!10}{}
\begin{center}
\resizebox{0.7\linewidth}{!}{
\begin{tabular}{lrrrrr}
\toprule 
\multirow{3}{*}{Parameter}  
& \multicolumn{5}{c}{Years}\\

\cmidrule(lr){2-6}
 & 2021 & 2022 & 2023 & 2024 & 2025 \\
\midrule 

Accepted Papers (Main Track)				& 1660	& 2064	& 2359	& 2719	& 2872 \\
PDFs Downloaded							& 1660	& 2071	& 2353	& 2713	& 2871 \\
URLs Extracted						& 1657	& 2069	& 2349	& 2711	& 2868 \\
LLM Detected Code URLs					& 960	& 1400	& 1558	& 1849	& 1962 \\
No URL in PDF					& 595	& 556	& 693	& 737	& 757 \\
Repos Cloned						& 737	& 1105	& 1240	& 1496	& 1559 \\
Repos Failed to Clone				& 223	& 295	& 318	& 353	& 403 \\
Train Code Available					& 618	& 927	& 1021	& 1175	& 1111 \\
Test Code Available						& 625	& 936	& 1041	& 1211	& 1233 \\
Train or Test Code Available						& 649	& 984	& 1079	& 1255	& 1267 \\
Model Weights Available						& 419	& 650	& 680	& 794	& 746 \\
Ack. Available							& 250	& 425	& 578	& 691	& 749 \\
Issues Related to Reproducibility					& 17	& 19	& 30	& 25	& 177 \\

\bottomrule 
\end{tabular}}
\end{center}
\end{table}

Finally, to understand which fields (topics) in the CV/ML domain are more active in open-sourcing their code, we classify each paper into a predefined set of topics. Specifically, we prompted the LLM using only the first two pages of the downloaded PDF and asked it to determine a single topic. 

The predefined topics (which were also LLM-generated to cover the 10 main topics of our five target conferences) are:
\begin{enumerate}[nosep]
    \item 3D Vision and Neural Rendering,
    \item Image and Video Synthesis,
    \item Multimodal Learning (Vision\,+\,Language\,+\,Reasoning\rlap{),}
    \item Large Language Models and Foundation Models,
    \item Generative AI and Diffusion Models,
    \item Reinforcement Learning and Decision Making,
    \item Trustworthy, Fair and Robust Machine Learning,
    \item Optimization and Theory of Learning,
    \item Scalability, Systems and Distributed ML,
    \item Application-Driven ML.
\end{enumerate}
In Prompt~\ref{prompt:classify_prompt}, we share the prompt used to classify each paper's topic.

We used InternLM~\cite{cai2024internlm2} (internlm3-8b-instruct) as the LLM. We finetuned the prompts until the provided results were identical to a human analysis based on a sample of 100 randomly selected papers from the latest editions of all venues.
In all prompts, we asked the LLM to also supply a reason and evidence for the final response. We observed that pushing the LLM to reason significantly improved the response quality and reduced the number of hallucinated responses. 

After optimizing our prompts on the 100 sampled papers, residual errors were marginal, with a slight tendency to overestimate the count of papers with code or weights. The actual situation of code and weight availability is thus actually slightly worse than what we report. (We do not have resources to get meaningful error bars for the analysis of these 55,377 papers, which would require manually creating a large pool of ground-truth statuses.)

We also extracted the number of citations for each accepted paper using the Semantic Scholar API~\cite{semantic_scholar} to further analyze the effect of code sharing on citations. (Forks and stars are not good metrics: some repositories without any code and just introducing a paper may have thousands of stars.)

\begin{table}[t]
\caption{Detailed statistics for ICCV.}
\label{tab:iccv_detailed_stats}
\rowcolors{4}{gray!10}{}
\begin{center}
\resizebox{0.7\linewidth}{!}{
\begin{tabular}{lrrrrr}
\toprule 
\multirow{3}{*}{Parameter}  
& \multicolumn{5}{c}{Years}\\

\cmidrule(lr){2-6}
 & 2021 & 2022 & 2023 & 2024 & 2025 \\
\midrule 

Accepted Papers (Main Track)			& 1617	& - 	& 2160		& - & 2701 \\
PDF Downloaded							& 1612	& - 	& 2156		& - & 2701 \\
URL Extracted							& 1610	& - 	& 2153		& - & 2698 \\
LLM Detected Code URL					& 1008	& - 	& 1508		& - & 1851 \\
No URL in PDF							& 520	& - 	& 556		& - & 752 \\
Repos Cloned							& 769   & -     & 1204		& - & 1455 \\
Repos Failed to Clone					& 239   & -     & 304		& - & 396 \\
Train Code Available					& 644   & -     & 994		& - & 946 \\
Test Code Available						& 659   & -     & 986		& - & 1071 \\
Train or Test Code Available						& 680	& - &	1035	& - &	1108 \\
Model Weights Available					& 454   & -     & 649		& - & 685 \\
Acknowledges Previous Repos				& 278   & -     & 542		& - & 653 \\
Issues Related to Reproducibility		& 50    & -     & 22		& - & 182 \\
\bottomrule 
\end{tabular}}
\end{center}
\end{table}

We present detailed results of this study for each venue in \cref{tab:cvpr_detailed_stats,tab:iccv_detailed_stats,tab:iclr_detailed_stats,tab:icml_detailed_stats,tab:nips_detailed_stats} for CVPR, ICCV, ICLR, ICML, and NeurIPS, respectively. 
There are some discrepancies between the number of accepted and downloaded papers, which is due to either scraping errors or, in times, the official numbers being incorrect, as in the case of CVPR'2022, where the CVF open access page indeed contains 2074 papers.
While the regular expression extracts all potential URLs, the LLM typically filters these to identify only the subset likely to represent the paper's actual implementation. Notably, the ``No URL in PDF" row reveals a significant number of instances where papers contain no URLs at all.
Unfortunately, not all the detected code URLs by the LLM, in fact, point to a valid code repository. Sometimes, the LLM detects non-code repositories, such as links to tools or software referred to in the paper, as possible code URL. However, most of the time, the code URL presented in the paper is not reachable, which reduces the number of cloned repositories.  

In some cases, even though a repository is successfully cloned to a local directory, it does not include any files except for a brief Readme file. This leads to a difference between the number of cloned repositories and the repositories with training and/or test code available. Another important observation is that a significant number of papers present only the training code but not the pretrained model weights, which is another barrier to reproducibility.   

We see the importance of open-sourcing the code in the ``Acknowledges Previous Repos'' row, which shows the number of repositories that acknowledge one or more publicly available repositories. As expected, this number increases each year for all the venues, notably doubling for ICLR from 2024 to 2025.
Finally, the number of issues regarding reproducibility in corresponding code pages is very low for open-sourced papers.

\begin{table}[t]
\caption{Detailed statistics for ICLR.}
\label{tab:iclr_detailed_stats}
\rowcolors{4}{gray!10}{}
\begin{center}
\resizebox{0.7\linewidth}{!}{
\begin{tabular}{lrrrrr}
\toprule 
\multirow{3}{*}{Parameter}  
& \multicolumn{5}{c}{Years}\\

\cmidrule(lr){2-6}
 & 2021 & 2022 & 2023 & 2024 & 2025 \\
\midrule 

Accepted Papers (Main Track)		& 860	& 1095	& 1584	& 2296	& 3827 \\
PDF Downloaded						& 860	& 1095	& 1574	& 2263	& 3708 \\
URL Extracted						& 860	& 1090	& 1568	& 2261	& 3703 \\
LLM Detected Code URL				& 695	& 918	& 1317	& 1945	& 3272 \\
No URL in PDF						& 117	& 128	& 181	& 228	& 261  \\
Repos Cloned						& 490   & 699   & 996	& 1464	& 2518 \\
Repos Failed to Clone				& 205   & 219	& 321	& 481	& 754  \\
Train Code Available				& 403   & 555   & 792	& 1116	& 1728 \\
Test Code Available					& 376   & 522   & 761	& 1105	& 1834 \\
Train or Test Code Available			
& 410	& 572	& 823	& 1199	& 1964 \\
Model Weights Available				& 166   & 236   & 341	& 496	& 837  \\
Acknowledges Previous Repos			& 79    & 161   & 243	& 397	& 720  \\
Issues Related to Reproducibility	& 16    & 27    & 13	& 20	& 60   \\

\bottomrule 
\end{tabular}}
\end{center}
\end{table}

We observe (\cref{fig:topics} that the proportion of theoretical papers (i.e., with topic “Optimization and Theory of Learning”, including “Statistical learning theory, optimization methods, generalization bounds, theory insights”) is decreasing while the proportion of theoretical papers with available code is generally increasing or stable. 
We thus concluded there was no surge of code-less theoretical papers, and thus that the observation of the decline or stagnation of papers with code was not biased by a reduction of the needs for empirical evaluation.

A similar code availability study is conducted in Paper2Code~\cite{seo2025paper2code}. However, since the authors only inspected the abstracts of the papers, their numbers differ from ours. Their inspection includes the accepted papers of the 2024 editions of ICLR, ICML, and NeurIPS, and they report that only 20\% of the papers have publicly available code, which is significantly lower than ours (around 50\%). Our study is more detailed, as we search for the code URLs in the entire paper (with a regexp).

\begin{table}[]
\caption{Detailed statistics for ICML.}
\label{tab:icml_detailed_stats}
\rowcolors{4}{gray!10}{}
\begin{center}
\resizebox{0.7\linewidth}{!}{
\begin{tabular}{lrrrrr}
\toprule 
\multirow{3}{*}{Parameter}  
& \multicolumn{5}{c}{Years}\\

\cmidrule(lr){2-6}
 & 2021 & 2022 & 2023 & 2024 & 2025 \\
\midrule 

Accepted Papers (Main Track)			& 1183	& 1233	& 1865	& 2634	& 3339\\
PDF Downloaded							& 1183	& 1233	& 1828	& 2610	& 3341\\
URL Extracted							& 1179	& 1232	& 1827	& 2606	& 3336\\
LLM Detected Code URL					& 763	& 883	& 1402	& 2058	& 2734\\
No URL in PDF							& 292   & 238	& 301	& 367	& 399\\
Repos Cloned							& 501   & 638   & 1023	& 1479	& 1947\\
Repos Failed to Clone					& 262   & 245	& 379	& 579	& 787\\
Train Code Available					& 354   & 481   & 746	& 1058	& 1317\\
Test Code Available						& 335   & 444   & 692	& 1046	& 1371\\
Train or Test Code Available			
& 370	& 495	& 779	& 1150	& 1485 \\
Model Weights Available					& 128   & 152   & 259	& 411	& 532\\
Acknowledges Previous Repos				& 63    & 116   & 206	& 373	& 486\\
Issues Related to Reproducibility		& 17    & 18    & 5	    & 16	& 27\\

\bottomrule 
\end{tabular}}
\end{center}
\end{table}

\begin{table}[t]
\caption{Detailed statistics for NeurIPS.}
\label{tab:nips_detailed_stats}
\rowcolors{4}{gray!10}{}
\begin{center}
\resizebox{0.7\linewidth}{!}{
\begin{tabular}{lrrrrr}
\toprule 
\multirow{3}{*}{Parameter}  
& \multicolumn{5}{c}{Years}\\

\cmidrule(lr){2-6}
 & 2021 & 2022 & 2023 & 2024 & 2025 \\
\midrule 

Accepted Papers (Main Track)		& 2334	& 2673	& 3218	& 4037	& 5290 \\
PDF Downloaded						& 2334	& 2671	& 3218	& 4035	& 5287 \\
URL Extracted						& 2327	& 2424	& 3202	& 4029	& 5280 \\
LLM Detected Code URL				& 1544	& 1674	& 2335	& 3479	& 5066 \\
No URL in PDF						& 637	& 576	& 684	& 2	    &   34 \\
Repos Cloned						& 1121	& 1270	& 1855	& 2621  & 2999 \\
Repos Failed to Clone				& 423	& 404	& 480	& 858   & 2067 \\
Train Code Available				& 860	& 977	& 1393	& 1903  & 1927 \\
Test Code Available					& 811	& 931	& 1349	& 1891  & 2079 \\
Train or Test Code Available			
& 897	& 1020	& 1491	& 2056 &	2608 \\
Model Weights Available				& 328	& 409	& 554	& 848   & 882  \\
Acknowledges Previous Repos			& 227	& 313	& 459	& 765   & 758  \\
Issues Related to Reproducibility	& 13	& 11	& 30	& 67    & 43   \\

\bottomrule 
\end{tabular}}
\end{center}
\end{table}

We attach all the code used in this study to the supplementary material. It will be made publicly available upon acceptance.

\begin{prompt}{Code URL Identification Prompt}
{Prompt template used for code URL identification.}
{prompt:url_prompt}

You will be given URL candidates extracted from a paper PDF, together with page snippets.

Your task is to identify which URL most likely points to the paper's open-source code:
\begin{itemize}
    \item GitHub repository
    \item GitLab repository
    \item code release page
\end{itemize}

Assume there is at most one true code repository for the paper.
All remaining URLs should be assigned to \texttt{other\_urls}.
A single evidence snippet is sufficient.

\textbf{Rules.}
\begin{itemize}
    \item Be strict.
    \item Include a URL only if the snippet suggests code or implementation, or if the URL itself is clearly a repository.
    \item Include at most the top-4 most relevant entries in \texttt{other\_urls}.
    \item Return only valid JSON.
    \item Do not return markdown or commentary.
\end{itemize}

\textbf{Output schema.}

\begin{verbatim}
{
  "code_urls": [
    {
      "url": "string",
      "confidence": 0.0,
      "reason": "short string",
      "evidence": [
        {
          "page": 0,
          "snippet": "string"
        }
      ]
    }
  ],
  "other_urls": [
    {
      "url": "string",
      "type": "dataset|project_page|paper|supplement|other",
      "reason": "short string",
      "evidence": [
        {
          "page": 0,
          "snippet": "string"
        }
      ]
    }
  ]
}
\end{verbatim}

\end{prompt}

\begin{prompt}{Repository Audit Prompt}
{Prompt template used for repository audit (training/inference availability).}
{prompt:audit_prompt}

You are auditing a GitHub repository for training and inference availability.

You must base every answer only on the provided repository evidence bundle:
\begin{itemize}
    \item README
    \item file tree
    \item embedded file contents
    \item optional issue excerpts
\end{itemize}

\textbf{Rules.}
\begin{itemize}
    \item Do not guess.
    \item If evidence is missing, set \texttt{value} to \texttt{null} and use:
    \texttt{unknown (not found in provided repo materials)}.
    \item Return only one valid JSON object.
    \item Do not return markdown or commentary.
\end{itemize}

\textbf{Evidence requirements.}
\begin{itemize}
    \item Every field must include an \texttt{evidence} list.
    \item Each evidence item must contain:
    \begin{itemize}
        \item \texttt{source\_type}: one of \texttt{README}, \texttt{FILE\_TREE}, \texttt{FILE\_CONTENT}, or \texttt{ISSUE}
        \item \texttt{location}: a precise pointer such as \texttt{README > Training}, a file path, or an issue identifier
        \item \texttt{quote}: an exact quote of at most 25 words
    \end{itemize}
\end{itemize}

\textbf{Conservative interpretation.}
\begin{itemize}
    \item \texttt{training\_code\_available = true} only if training-related code or explicit training instructions are present.
    \item \texttt{inference\_or\_test\_code\_available = true} only if inference/evaluation code or explicit instructions are present.
    \item \texttt{weights\_available\_for\_this\_method = true} only if checkpoints are provided for the audited method.
    \item \texttt{presents\_dataset = true} only if the repository itself presents a dataset.
    \item \texttt{open\_issues = true} only if issues explicitly mention installation problems, missing steps, reproducibility, or similar concerns.
    \item Ignore issues authored by \texttt{NielsRogge}.
    \item Include the full repository URL.
\end{itemize}

\textbf{Output schema.}

\begin{verbatim}
{
  "repo": {
    "url": "",
    "name": ""
  },
  "assessment": {
    "is_training_based": {
      "value": null,
      "rationale": "",
      "evidence": []
    },
    "requires_specific_training": {
      "value": null,
      "rationale": "",
      "evidence": []
    },
    "training_code_available": {
      "value": null,
      "rationale": "",
      "evidence": []
    },
    "inference_or_test_code_available": {
      "value": null,
      "rationale": "",
      "evidence": []
    },
    "training_instructions_available": {
      "value": null,
      "rationale": "",
      "evidence": []
    },
    "inference_or_test_instructions_available": {
      "value": null,
      "rationale": "",
      "evidence": []
    },
    "weights_available_for_this_method": {
      "value": null,
      "rationale": "",
      "evidence": []
    },
    "presents_dataset": {
      "value": null,
      "rationale": "",
      "evidence": []
    },
    "open_issues": {
      "value": null,
      "rationale": "",
      "evidence": []
    },
    "acknowledgement": {
      "value": null,
      "rationale": "",
      "evidence": []
    }
  },
  "open_issues_lowerbound_notes": {
    "value": "",
    "evidence": []
  }
}
\end{verbatim}

\end{prompt}

\begin{prompt}{Paper Topic Classification Prompt}
{Prompt template used to classify the topic of each paper.}
{prompt:classify_prompt}

You are classifying research paper PDFs into exactly one predefined topic.

You will be given a bundle of papers. 

For each paper:
\begin{itemize}
    \item choose the single best topic
    \item justify it using evidence from the provided PDF snippets only
\end{itemize}

\textbf{Rules.}
\begin{itemize}
    \item Do not guess.
    \item Use only the provided PDF snippets.
    \item Do not use outside knowledge.
    \item Each paper must have exactly one selected topic, or \texttt{null} if unknown.
    \item If the paper cannot be confidently assigned, set \texttt{topic\_id} to \texttt{null} and use exactly:
    
    \texttt{unknown (insufficient evidence in provided PDF snippets)}
    
    \item Return only valid JSON.
    \item Do not return markdown or commentary.
\end{itemize}

\textbf{Evidence requirements.}
\begin{itemize}
    \item Every paper entry must include an \texttt{evidence} list.
    \item Each evidence item must contain:
    \begin{itemize}
        \item \texttt{page}: integer page index
        \item \texttt{snippet}: exact snippet copied from the provided materials
    \end{itemize}
    \item Keep each snippet at most 25 words.
    \item Do not include double quotes in the snippet field.
    \item Remove or avoid unparsable characters such as math symbols or broken glyphs.
    \item Keep the rationale short, ideally 1--2 sentences.
\end{itemize}

\textbf{Topics.}
\begin{enumerate}
    \item 3D Vision and Neural Rendering
    \item Image and Video Synthesis
    \item Multimodal Learning (Vision + Language + Reasoning)
    \item Large Language Models and Foundation Models
    \item Generative AI and Diffusion Models
    \item Reinforcement Learning and Decision Making
    \item Trustworthy, Fair and Robust Machine Learning
    \item Optimization and Theory of Learning
    \item Scalability, Systems and Distributed ML
    \item Application-Driven ML
\end{enumerate}

\textbf{Output schema.}

\begin{verbatim}
{
  "results": [
    {
      "paper": {
        "paper_id": "string",
        "filename": "string"
      },
      "topic": {
        "topic_id": 0,
        "topic_name": "string"
      },
      "confidence": 0.0,
      "rationale": "string",
      "evidence": [
        {
          "page": 0,
          "snippet": "string"
        }
      ]
    }
  ]
}
\end{verbatim}
\end{prompt}

\section{Details and examples regarding the verification framework}
\label{sec:detailverifframework}

\subsection{Log verification details and examples}
\label{app:examplelog}

If the authors trained a model, they most certainly already have logs, from which they visually monitored training and execution curves while developing their approach. The extra work to provide log verification material for reviewers is then just to document log information.

Besides snapshots of log curves pointing at the corresponding experiments in the paper (e.g., a line in a SOTA table), log verification material should include a common core of information for both training and test. This information includes the following fields:
\begin{itemize}[nosep]
    \item a clear \field{reference to an experiment} in the paper, e.g., a line in a SOTA table,
    \item the \field{dataset size} used in the run,
    \item \field{GPU utilization} and \field{CPU utilization},
    \item \field{model size} (number of parameters),
    \item \field{FLOPs} (or a comparable compute estimate).
\end{itemize}
On top of these shared fields, training logs must report:
\begin{itemize}[nosep]
    \item the \field{learning scheme} and \field{training parameters},
    \item the \field{loss values} over steps or epochs,
\end{itemize}
while validation logs must report
\begin{itemize}[nosep]
    \item the final \field{evaluation metrics}.
\end{itemize}
including what was actually reported in the paper.
In the Appendix, we present example log snapshots taken from TensorBoard, MLflow, and W\&B with the required fields.

\begin{figure*}[t]
\centering
\begin{subfigure}[t]{1.0\linewidth}\centering
  \includegraphics[width=\linewidth]{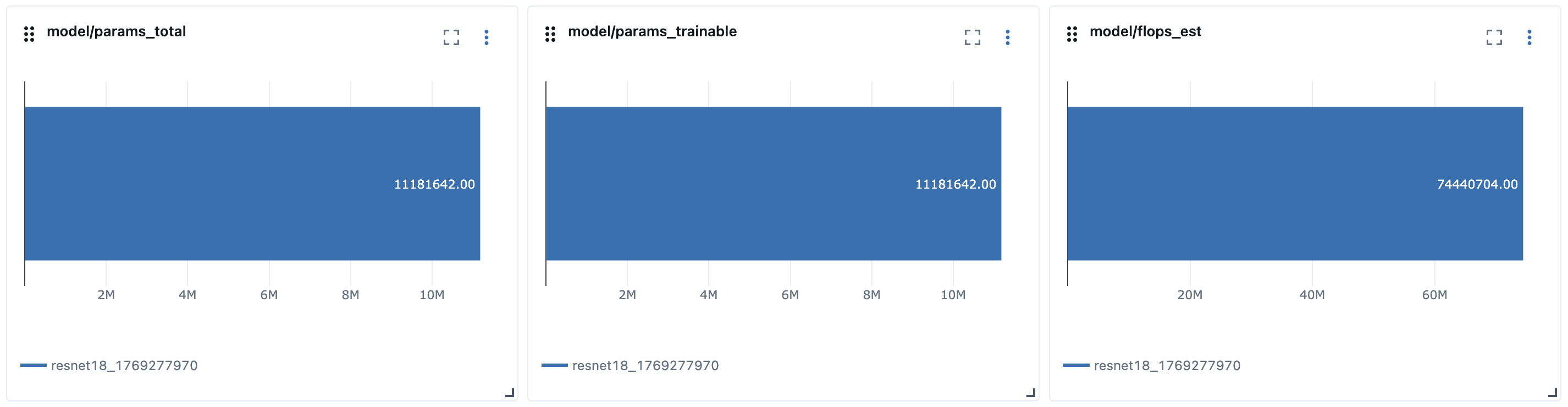}
\end{subfigure}\hfill%
\begin{subfigure}[t]{1.0\linewidth}\centering
  \includegraphics[width=\linewidth]{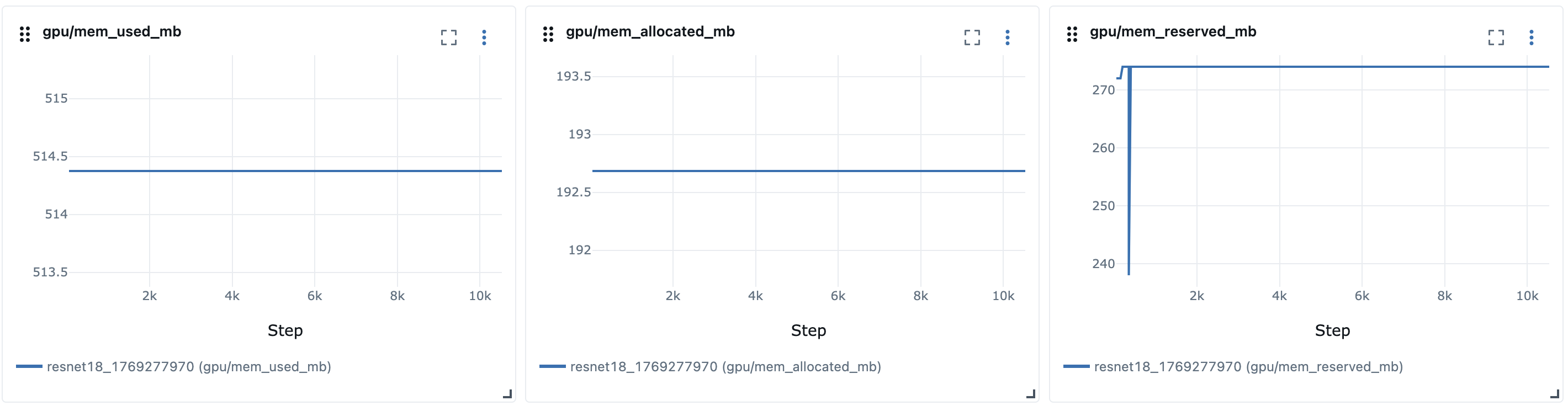}
\end{subfigure}\hfill%
\begin{subfigure}[t]{1.0\linewidth}\centering
  \includegraphics[width=\linewidth]{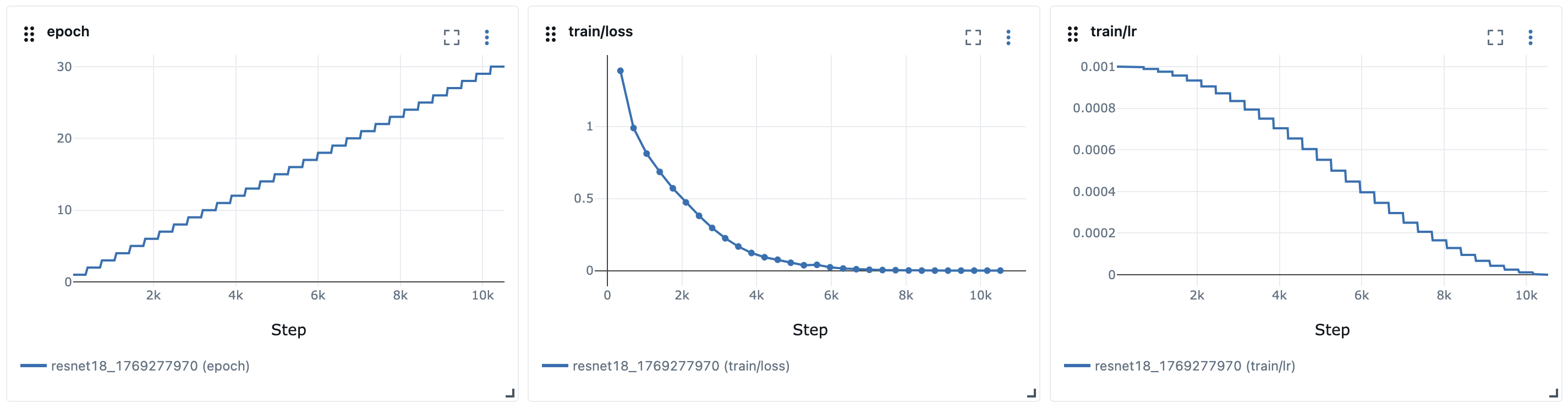}
\end{subfigure}\hfill%
\begin{subfigure}[t]{1.0\linewidth}\centering
  \includegraphics[width=\linewidth]{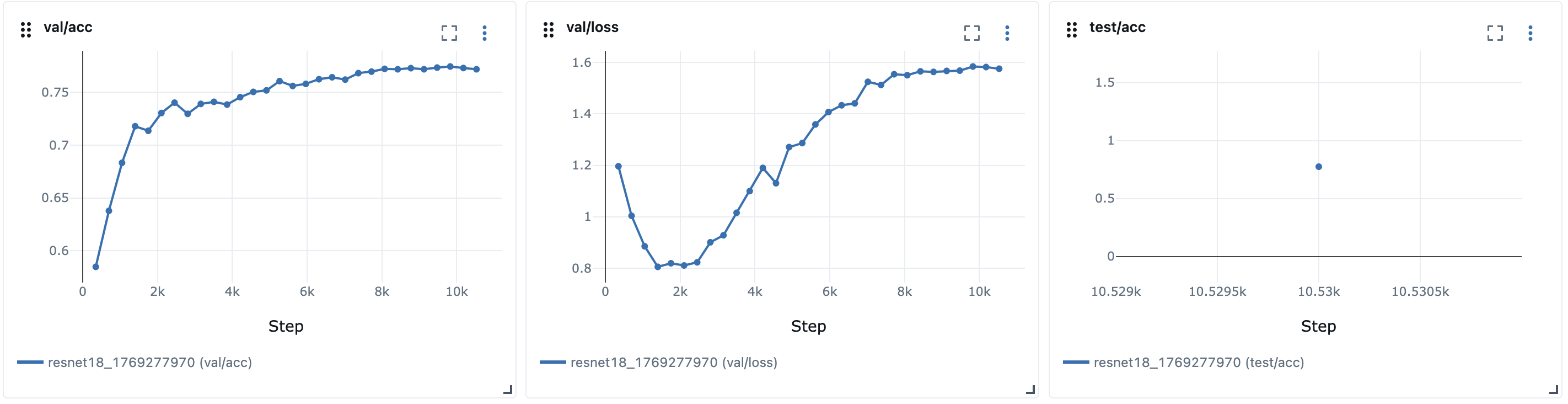}
\end{subfigure}

\caption{Example log plots from MLFlow.}
\label{fig:mlflow_logs}
\end{figure*}

\begin{figure*}[t]
\centering
\begin{subfigure}[t]{1.0\linewidth}\centering
  \includegraphics[width=\linewidth]{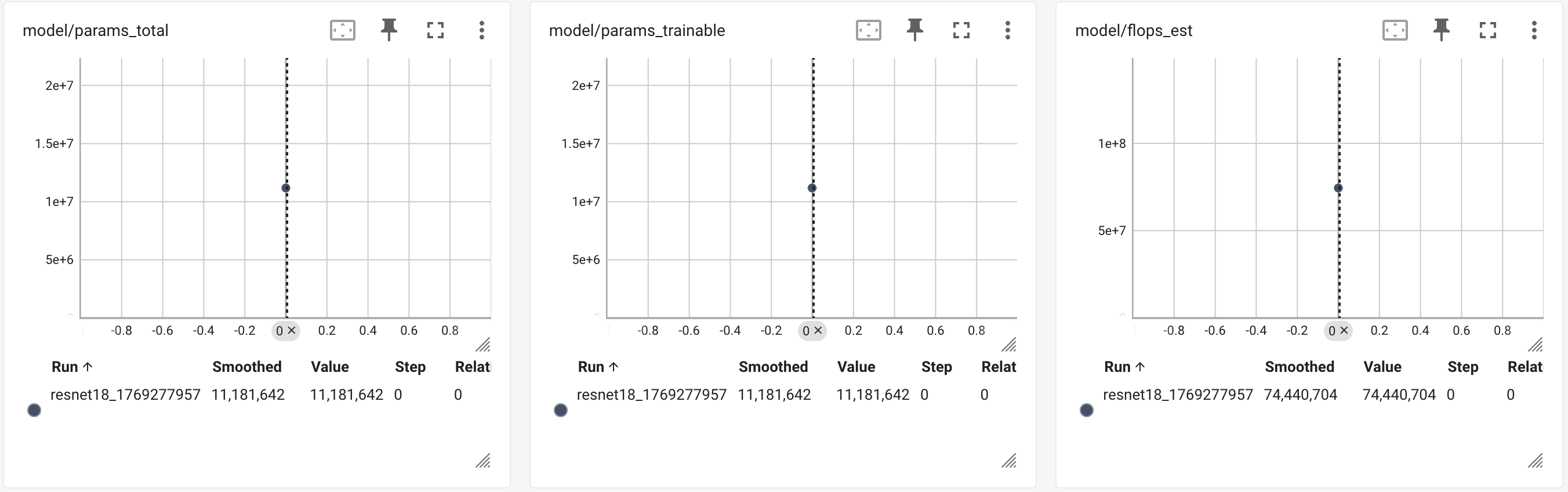}
\end{subfigure}\hfill%
\begin{subfigure}[t]{1.0\linewidth}\centering
  \includegraphics[width=\linewidth]{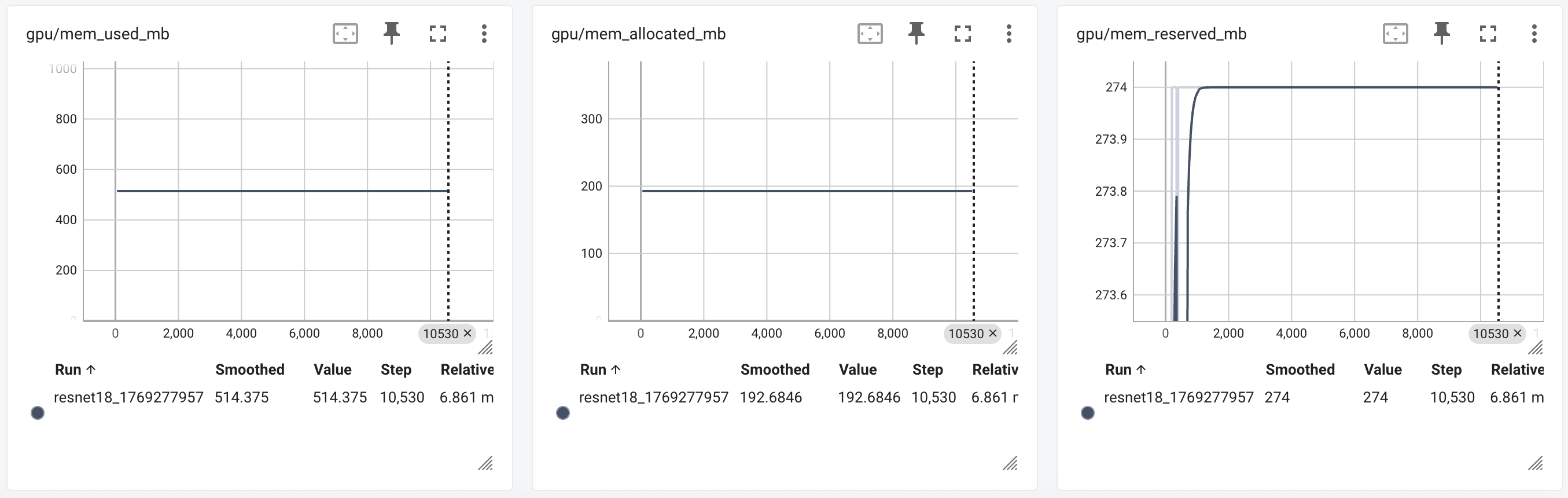}
\end{subfigure}\hfill%
\begin{subfigure}[t]{1.0\linewidth}\centering
  \includegraphics[width=\linewidth]{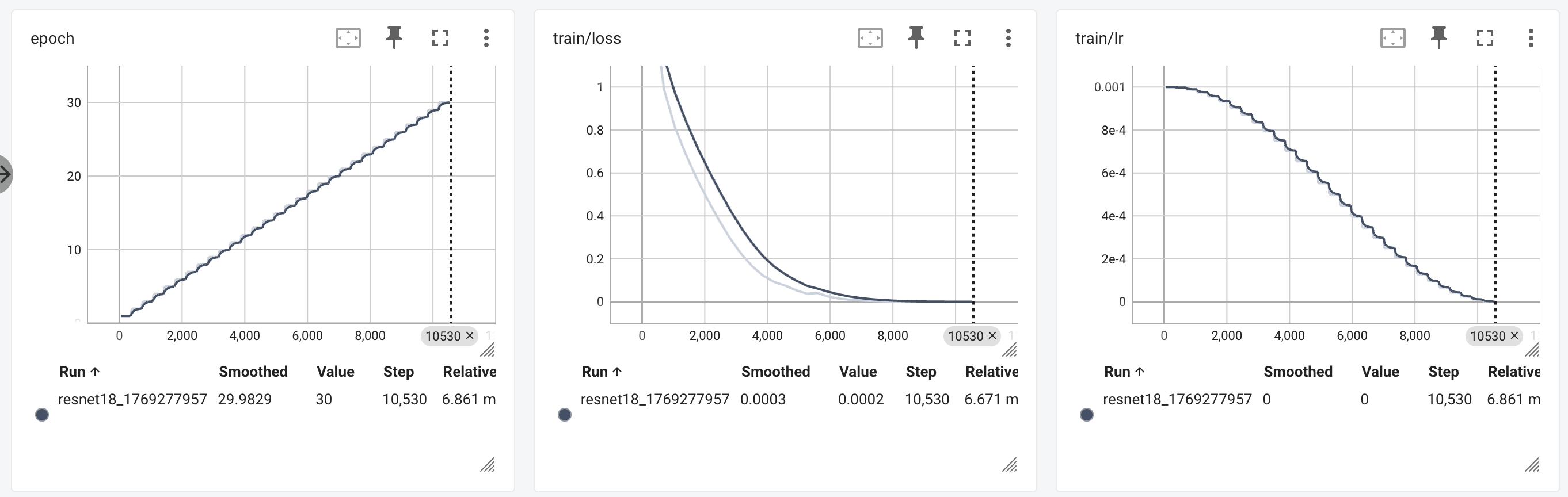}
\end{subfigure}\hfill%
\begin{subfigure}[t]{1.0\linewidth}\centering
  \includegraphics[width=\linewidth]{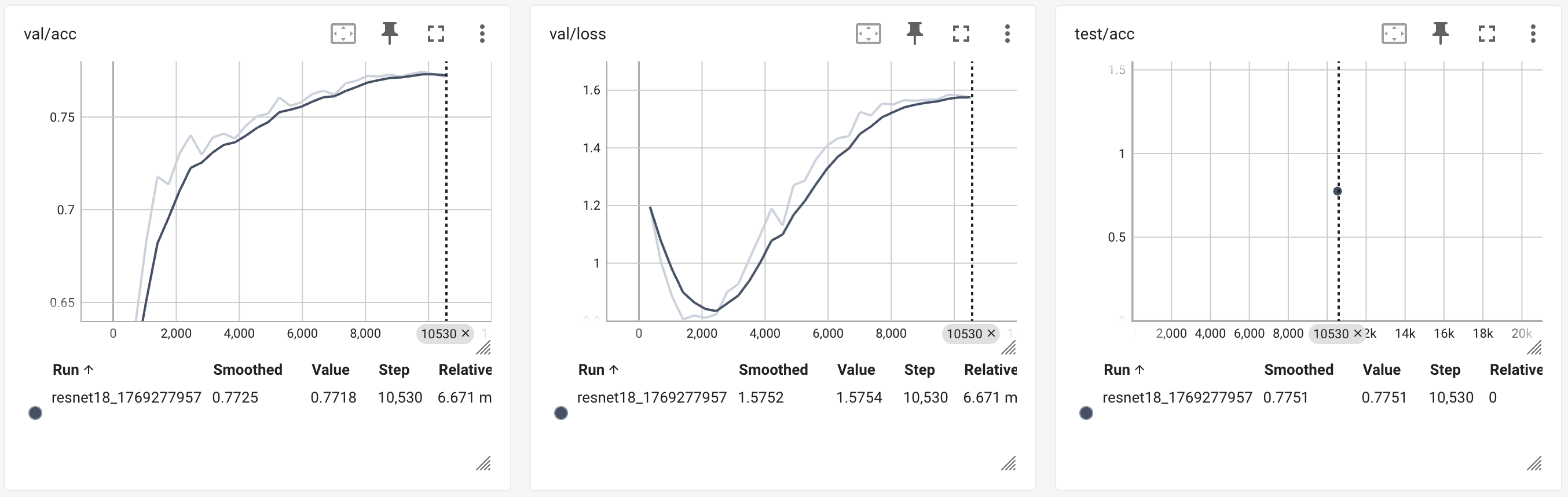}
\end{subfigure}

\caption{Example log plots from TensorBoard.}
\label{fig:tb_logs}
\end{figure*}

\begin{figure*}[t]
\centering
\begin{subfigure}[t]{1.0\linewidth}\centering
  \includegraphics[width=\linewidth]{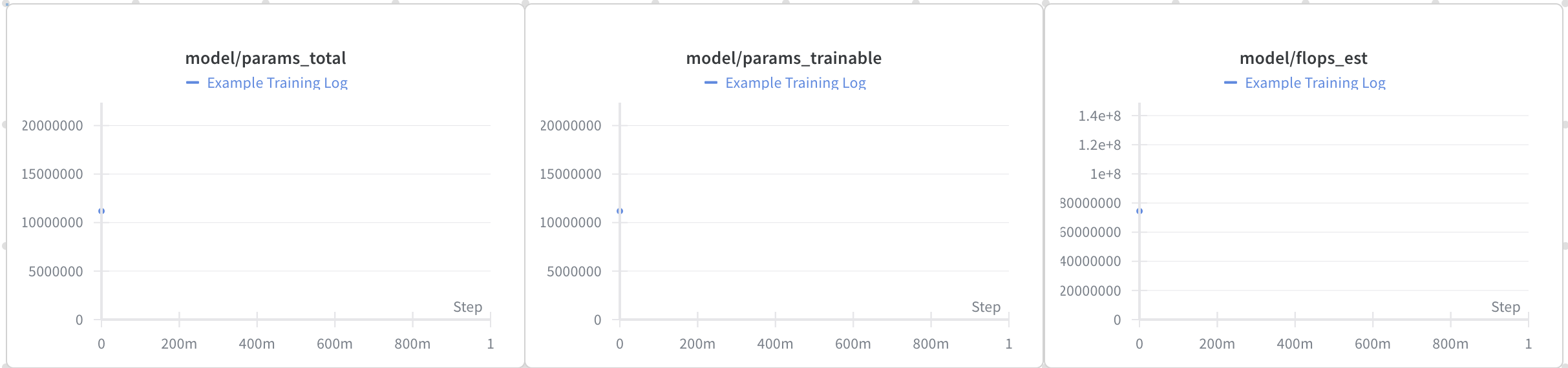}
\end{subfigure}\hfill%
\begin{subfigure}[t]{1.0\linewidth}\centering
  \includegraphics[width=\linewidth]{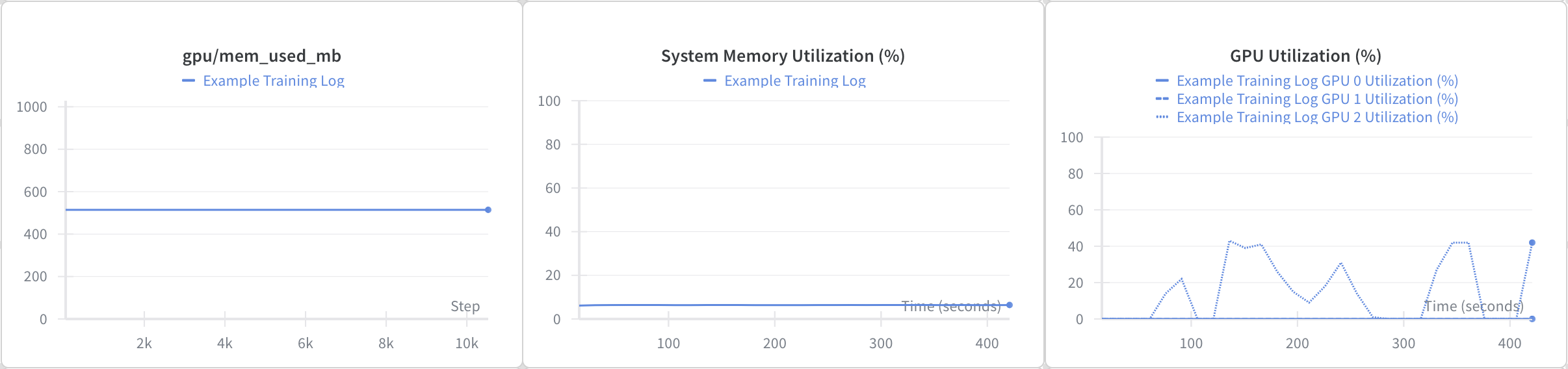}
\end{subfigure}\hfill%
\begin{subfigure}[t]{1.0\linewidth}\centering
  \includegraphics[width=\linewidth]{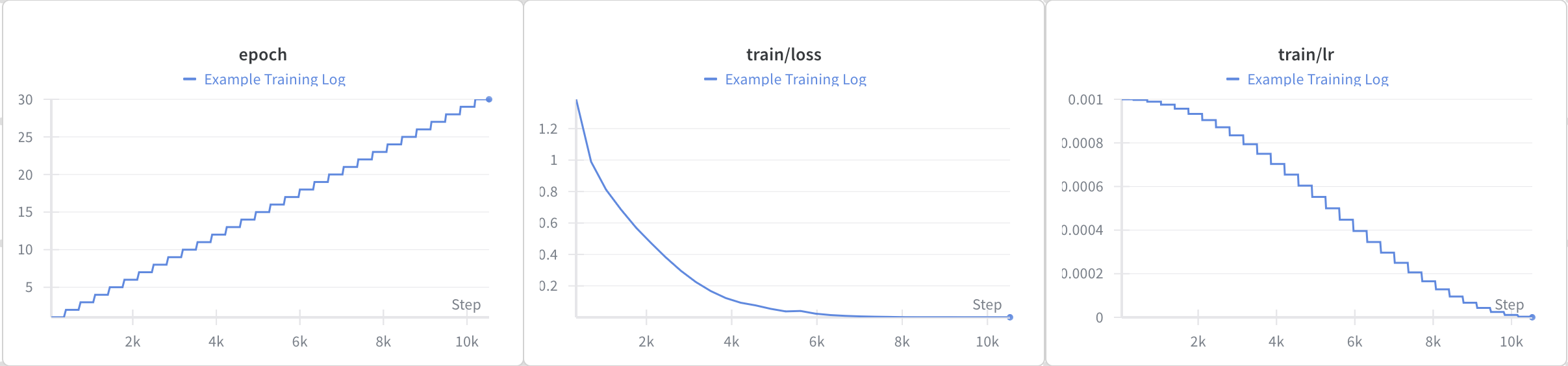}
\end{subfigure}\hfill%
\begin{subfigure}[t]{1.0\linewidth}\centering
  \includegraphics[width=\linewidth]{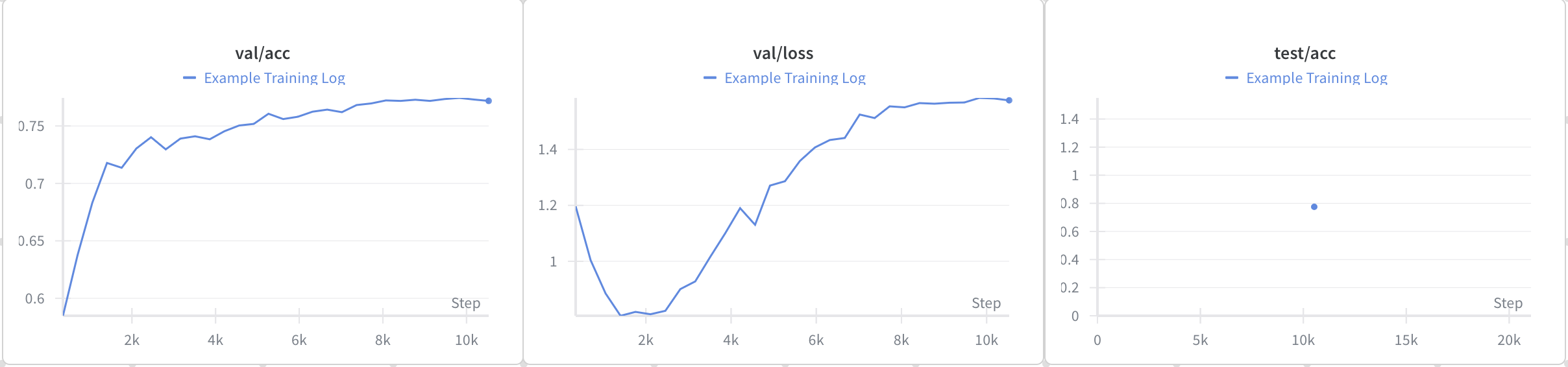}
\end{subfigure}

\caption{Example log plots from W\&B.}
\label{fig:wandb_logs}
\end{figure*}

As examples of log verification material, we show representative visualizations produced by three widely adopted experiment-tracking tools for custom training: MLflow (Figure~\ref{fig:mlflow_logs}), TensorBoard (Figure~\ref{fig:tb_logs}), and Weights \& Biases (Figure~\ref{fig:wandb_logs}).
To facilitate comparison, we organize the plots into four thematic groups that roughly follow the life cycle of a training run.

The first group summarises \textit{model characteristics}, including the total and trainable parameter counts, as well as an estimate of computational cost (FLOPs). The second group reports \textit{system-level signals} captured during training, such as GPU memory usage (used, allocated, and reserved). When supported by the logging backend, we additionally track host memory utilization and GPU utilization. The third group focuses on \textit{optimization procedure}, recording the number of epochs completed alongside training loss and the learning-rate schedule. Finally, the fourth group presents \textit{evaluation metrics} on the validation set and, when available, the test set, together with per-epoch validation loss.

For each experiment to verify, the task (P2) of the reviewer includes, but is not limited to:
\begin{itemize}[itemsep=1pt,topsep=2pt]
\item checking that the training process is consistent with the paper,
\item assessing training convergence, including performance variance,
\item checking the performance consistency with the paper,
\item detecting bad practices, e.g., training on more data that said (e.g., comparing the number of epochs vs the batch size and iteration count) or cherry-picking checkpoints.
\end{itemize}

\subsection{Metric verification details}
\label{app:metricverif}

As mentioned in the main paper, software (code and data) is central to reproducibility for computer science papers. (Note however that computer science papers sometimes also include experiments that are not purely virtual, such as robotic manipulations. They may include human studies too, e.g., to evaluate acceptability, assess image realism, or measure the time in a computer-assisted human labeling process. We only consider here repeatable \emph{virtual} experiments with mathematically-defined metrics.)

If the authors reported some metrics, then they already processed generated outputs using some evaluation code and, possibly, ground-truth data. While training or inference code could be sensitive for the authors, metric code should not, or it should at least be accessible to reviewers, who already currently commit to treat all information related to submissions as confidential. In fact, a metric must anyway have a public definition that enables its implementation. We presume that, most of the time, the evaluation code and data will be standard and readily available.

The extra task then just amounts to:
\begin{itemize}[itemsep=1pt,topsep=2pt]
    \item packaging the output files and ground-truth data (if any), 
    \item making the evaluation code stand-alone, if not already the case.
\end{itemize}
A code assistant can typically see to it.

To control if errors were introduced when computing metrics or when reporting them, the verification material of an experiment should include:

\begin{itemize}[
     itemsep=1pt,
     topsep=2pt,
]
    \item a \field{definition} or \field{reference} to the experiment in the paper, 
    \item a \field{JSON file} representing the output of the experiment,
    \item a link to the \field{metric evaluation code} from an official dataset or benchmark, or if the metric is original, the anonymized evaluation code, which may include a trained model (e.g., to propose alternative features to an FID-like evaluation \cite{heusel2017fid}),
    \item a link to \field{ground-truth data} used for the evaluation, whether public or proprietary if the test data is original, 
    \item a \field{guide} for download and usage.
\end{itemize}
No license agreement should be needed to access proprietary data because it could break the anonymity of the authors or of the reviewer.

To prevent installation issues, as reviewers have to check the evaluation material and rerun it, we actually recommend that authors prepare an anonymized Google Colab-like notebook containing:
\begin{itemize}[itemsep=1pt,topsep=2pt]
    \item \textit{output data}, or code to download them anonymously, \item \textit{ground-truth data} or, preferably, code to download them from standard repositories, e.g., original dataset web site or Hugging Face datasets,
    \item \textit{evaluation code}, or code to download it, using standard procedures when applicable, e.g., calls to PyTorch libraries or dataset devkits.
\end{itemize}
Examples of such notebooks are available from \url{https://github.com/papersubmissions13/PositionPaper}.

Checking the metric evaluation software includes, but is not limited to:
\begin{itemize}[itemsep=1pt,topsep=2pt]
    \item checking the genuineness of provided or downloadable ground-truth data, e.g., making sure difficult samples were not excluded,
    \item checking the validity of provided or downloadable evaluation code,
    \item running the evaluation code and checking the result consistency with the paper,
    \item reporting on it in the review and balancing it for the recommendation (P4).
\end{itemize}

The implied discipline to provide and review such material is also expected to help the community standardize evaluation codes, contributing to more fair evaluations and improved paper quality.

\subsection{Verification process}

\paragraph{A priori verification vs a posteriori replication.}

The target of our position paper is not science in general but, more modestly, the organization of ML\,/\,CV conferences. It is the responsibility of program chairs to select papers (1)~that are sound and (2)~that have potential to make a significant impact in their field. We believe a systematic a priori verification at review time, even if it is partial and possibly sidestepped, would be a notable improvement towards more soundness and reproducibility. In contrast, a posteriori code verification is highly fortuitous (limited code availability, need for good-will users, substantial effort, resource requirements) and comes too late, after acceptance, in a publication landscape which is certainly not without errors but that chiefly ignores retraction.

In fact, our original motivation is to address “toxic” papers, with partly hidden experimental protocols and irreproducible results. We believe that such papers, which prevent meaningful other papers from being submitted or accepted because they are not SOTA, would have a harder time being published if some basic verifications could be done at review time to find traces of secret recipes.

Our hope, if such a verification becomes a standard, is that it will actually contribute to filtering out papers from careless authors, thus promoting better scientific practices. However, it will not prevent authors that deliberately forge results from also forging verification material. Totally preventing forgery would not only require full code and data availability (hence raising proprietary issues), but also demand a crazy amount of verification effort from reviewers, including rerunning training and inferences (assuming reviewers have the time and resources to do so).

\paragraph{Verification burden.}

For common cases, verifying both log data and evaluation code should only be a matter of minutes (not counting execution time). Let's considering the short evaluation script example that we provide, It is 35 lines long but less than half of them actually matter (disregarding empty lines, comments, title printing, file loading check, and import commands). As they contain little or no algorithmic content, checking the logic of these lines actually requires less than 5 min. Arguably, as a comparison, this is less difficult and takes less time than checking a formal proof in a paper. For authors, typesetting a proof also takes longer than writing such a code. 

Assuming that most papers have 1-4 main empirical claims regarding 1-3 metrics, with some sharing regarding data and metrics, and given that reviewers are often familiar with the dataset and metrics, we consider the total verification time to be less than one hour in general, and often much less. It is a price to pay, but it is particularly small compared to a full-fledged code verification.

Area and program chairs only have to take into account extra comments regarding verification when assessing a paper (P3, P5). And the extra load after acceptance (P6-8) is very light for all actors.
In any case, improving verifiability cannot come for free. Our verification effort however is minor for all actors, offering a good compromise regarding enhanced guarantees.

\paragraph{Overhead variance.}

Just like some papers are much easier than others to read and review, we expect some variance in the verification effort. While multiple datasets further increase the overheads, limiting the verification to SOTA claims restrains the required effort, including for large benchmark papers. As for heavy outputs, we expect most of the complexity to be hidden in JSON files and, most often, in standard evaluation procedures. To reduce the overall effort, we also suggest a single metric reviewer (see below).

\paragraph{Single evaluator.} 

To keep the workload low, we propose that there be a single metric evaluator per paper. This substantially reduces the workload of reviewers on average. For NeurIPS last year (4-6 papers per reviewer and 4-5 reviewers per paper), metric evaluation would have concerned 1 paper per reviewer, occasionally 2. And it is overestimated as it assumes that all submissions provide evaluation code, whereas there is actually a number of papers for which such an evaluation either does not apply (e.g., theoretical papers) or is not possible (see Exemptions in \cref{sec:policy}) --- which is ok. A possible lighter alternative, to be discussed with the community, is that the reviewer assigned to log and metric verifications reads the paper to understand experiments but does not review it in depth, focusing only on experiment verification.

\paragraph{Escaping verification and forged verification material.}

Verification can be escaped with proper justification (Exemptions in \cref{sec:policy}). Still, we think it is better than nothing. We also hope the authors can catch some errors themselves while preparing verification material, which they could have missed otherwise. Besides, forged verification material constitutes evidence of misconduct. If offenders are caught, it will be harder for them to pretend it was a lapse in attention.

\section{Code before and after acceptance}
\label{app:code}

\subsection{Code as full-fledged part of the reviewing}
\label{app:codereviewed}

\paragraph{Reviewing code.}

The safest but most radical option to try to enforce reproducibility is to impose code reviewing. Although it would not address all possible reasons for not disclosing code, the authors providing code could get the formal guarantee that reviewers will undertake not to uncover any information from the submitted material.

It opens the possibility for reviewers to fully check the code attached to an experiment and rerun it. However, not all experiments could be reproduced in this way, particularly those requiring a lot of resources (compute, memory space, time, etc.) Besides, it requires a lot of effort from reviewers.

Ideally, validation would involve uploading code, models, and data to an independent third-party server, secured to allow uploads of proprietary software.
The server would run the code and automatically generate a report, while an LLM-based auditor would inspect the code for suspicious behavior, such as hard-coded outputs, lookup tables, or other shortcuts that could artificially reproduce the claimed numbers. Unfortunately, this framework is unlikely to be feasible in the near future because hosting, securing, and maintaining such a platform is expensive and operationally complex.

\paragraph{A few authors already submit code.}

We do not have large-scale statistics regarding authors that choose to provide code as supplementary material. We only observe, as ACs for recent CV conferences (CVPR 2025, 2026 and ECCV 2024, 2026), that we only got 16/74 submissions with some code, i.e., a bit more than 20\%. And out of the 48 reviews for the 16 submissions with code, only two reviewers mentioned the code, and just to say they “appreciate the effort toward reproducibility”. Besides, one reviewer of a codeless submission complained that code was not submitted ``for verification''. Last, a number of reviewers asked for the code release plan, thus implicitly trusting the authors.

However incomplete and topically biased this experience may be, we consider that only a small minority of submissions include code, and that it is hardly ever used in the review process. This is understandable given the effort that is required to actually review code; the best that can be expected is to allow reviewers to possibly peek at the code when something is unclear in the paper description.

\subsection{Code after acceptance}
\label{app:codeaccepted}

\paragraph{The ``code being polished'' excuse.}

The argument that ``time is needed to polish the code'' (\cref{sec:codenotavail}) does not hold much nowadays, as cleaning up and reorganizing code can largely be done by code assistants. AI indeed makes code preparation easier, but it is not a magic wand. Most of the time spent to make code available is not in polishing it or even documenting its use, but in making sure it can be installed, with relevant data, and rerun with results that are consistent with the paper. Code polishing sounds more of an excuse to delay availability than a real reason.

\paragraph{Discussing the code virtue hypothesis.}

The code virtue hypothesis is to consider that, by making code available, authors get their own reward: if they provide good code, it will be reused, their paper will be cited more (which is good for them) and science will advance (which is good for humanity).

Now if, after acceptance, authors provide no code, or only partial code, or code/weights that does not succeed in reproducing results in the paper, or weights that do reproduce good results but that do not generalize because they were trained on test sets, they will nevertheless be cited (which is good for them) but it will not advance science; on the contrary, it will prevent other papers from being published or cited because they will not be SOTA (which is unfair and bad for humanity).

In fact, if providing good code was enough of an incentive to get more citations, it would have become the standard and we would not witness a plateauing in the proportion of code made available (40-50\%) nor an increase in open issues related to reproducibility in code repositories of accepted papers in the last 3 years: 1.9\% in 2023, 2.3\% in 2024, 4.0\% in 2025 (numbers computed from \cref{tab:cvpr_detailed_stats,tab:iccv_detailed_stats,tab:iclr_detailed_stats,tab:icml_detailed_stats,tab:nips_detailed_stats}). The proportion goes even up to 4.9\% in 2025 for papers that make both train and test code available.

We believe that the increasing pace of science and the pressure to publish more undermines a behavior that otherwise should indeed be virtuous. We are not saying that all authors of papers with reproducibility issues are intentional fraudsters; some probably are, but others are simply the victims of carelessness or inadvertent errors.

Rather than let time decide on the soundness and significance of a paper, our proposal is to include an extra verification dimension at review time to try to filter out some of these problematic papers \textit{before they are published}. Because afterwards, it is too late \cite{nag2025global}.

Related to this hypothesis, our citation analysis (\cref{fig:citations}), which observes that papers with code have twice as many citations as paper without code, is correlational and not causal. Nevertheless, we consider that a large number of people think that available code helps increase the number of citations. This belief should be enough as an incentive to provide code, but it seems to be reaching its limits (\cref{sec:studycode}). In any case, this motivating assumption does not invalidate our proposition for an a priori verification.

\paragraph{Limitations.}

While well-captioned training curves can reveal cherry-picked checkpoints, and training on more data than said, it cannot catch all forms of data leakage and other methodological flaws, nor lookup cheating and forged logs or output files. 
In fact, the verification we propose is only partial, and can be fooled by forged verification data, which is easily done nowadays with an AI. It is an accepted limitation.

\end{document}